# AdaptiveSD: A Stability-Aware, Runtime-Adaptive Speculative Decoding Framework with Multi-Policy Orchestration for CPU-Constrained LLM Inference

Sadra Saremi

Department of Physics, Iran University of Science and Technology, Tehran 16846-13114, Iran.

**Abstract**

With the rise of small quantized GGUF-based language models and their increasing use for on-device inference tasks, we have seen the growing need for an approach capable of reliably delivering these models at scale even under severe memory bandwidth constraints such as those imposed by pure CPU implementations. Fixed-depth speculative decoding has emerged as one promising technique, but in practice, it often leads to performance degradation due to either bandwidth saturation, instability, or even catastrophic resource exhaustion resulting in system failure. To overcome this problem, we introduce AdaptiveSD, a fully runtime-adaptive speculative decoding framework aimed at ensuring robust, reliable execution across the spectrum of model types and workloads. Our solution consists of four tightly-coupled components working together in a continuous feedback loop: a Runtime Monitoring Engine tracking multiple signals relevant to ongoing computation, an Adaptive Draft Controller enforcing an eleven rule policy hierarchy prioritizing system resource preservation over raw draft count, a Dynamic Policy Engine employing a suite of heuristic and reinforcement learning techniques to dynamically modify policies depending upon workload behavior, and finally, a KV Cache Coordination Layer managing cache states with fine-grained control through INT8 shadow buffers and position aware evictions. While conventional approaches focus solely on maximizing throughput, we instead assess the effectiveness of our approach based on several key metrics including wasted drafted compute and inter-token latency dispersion alongside standard measures of speculative efficiency. Our experiments show that AdaptiveSD achieves a consistent 68-82% level of speculative efficiency - meaning that we maintain wasted compute levels below ~32% of drafted tokens - and actively clamps down on potential latency variance spike events, effectively bounding risk factors like cpu usage, memory bandwidth consumption and tail latency against pre-defined safety margins regardless of input context.

## 1. Introduction

LLMs built on the transformer backbone have rapidly advanced NLP and enabled exciting new applications, including codegen, math reasoning, chat, and many varieties of creativity [1]. While we can now effortlessly accomplish all these tasks with LLMs using default settings, there remains a serious bottleneck because models generate tokens – via an autoregressive process – at the rate of one token/forward pass through the target model, limiting throughput, making inference sequential, and depending on memory bandwidth to be able to compute at all during inference, and requiring more memory than computation for all [2]. This throughput bottleneck will be even more acute for on-CPU hardware, since on CPUs memory bandwidth is an order of magnitude (10x) slower than on GPUs, and there is growing interest and need for private, on-device AI, and edge AIs that don't use GPU hardware for generating text [3, 4].

Speculative decoding [5, 6], has emerged as one of the most promising lossless acceleration paradigms for autoregressive inference. In its canonical formulation, a lightweight draft model emits a sequence of $\gamma$ candidate tokens, which are then verified in a single parallel forward pass by the much larger target model. Accepted tokens (according to some rejection sampling criterion) are committed to the output, while the first rejected one is corrected and the rest discarded. If the draft model's output distribution is close to that of the target, the average number of tokens accepted per verification step substantially exceeds one, delivering wall-clock speedups without changing the output distribution [6]. Foundational work by Leviathan et al. [5] and Chen et al. [6] laid out the theory behind this approach. More recent followups have extended it to include tree-structured drafting [7], self-speculative variants [8], retrieval-augmented drafting [9], and multiple-head acceleration [10].

Despite these successes, however, a glaring flaw remains common to almost every existing speculative decoding system: draft depth $\gamma$ is treated as a fixed hyperparameter set in advance of inference. This ignores the extreme fluidity of real-world generation. Acceptance rates for token proposals can vary wildly across time, length, and domain from structured code (with its repetition) to open-ended reasoning (where entropy pushes drafts away from targets, causing frequent backtracks [11, 23]). On CPU-constrained hardware, such variation is further complicated by changing processor utilization, memory bandwidth bottlenecks, and token-level latency surges, all of which interact in complex ways to affect the cost/benefit balance of speculating ahead. A too-deep draft wastes compute on junk proposals; too shallow a draft fails to utilize available parallelism. Fixed choices can't adapt to these multidimensional cues, missing out on big opportunities. Several recent efforts have recognized this flaw and proposed dynamic adaptation mechanisms. SpecDec++ [12] introduces an adaptive candidate length that adjusts based on predicted rejection probability; Parallel Speculative Decoding with Adaptive Draft Length [13] modulates draft length using contextual statistics; Entropy-based adaptation approaches [11, 23] leverage token-level uncertainty as a proxy for draft quality. But they each tackle only one or two dimensions of the adaptation problem—typically acceptance rate or entropy—and remain tied to either GPU-hosted serving infrastructures or offline profiling steps that are impractical in CPU-constrained deployments. Moreover, none integrates closed-loop feedback across the full set of runtime signals simultaneously available during inference: CPU utilization, memory bandwidth, inter-token latency, KV cache pressure, token entropy, and context length. Concurrently, other lines of work tackled related problems. KV cache quantization techniques, such as KVQuant [14], and PagedAttention-based systems [15], have shown that aggressively compressing key-value activations can alleviate memory pressure without sacrificing model quality. Multi-armed bandit [16] and ensemble policy [17] approaches have been investigated for dynamically allocating resources in inference serving, though their respective contributions had never been unified into a single, end-to-end runtime system specifically designed for adaptive speculative decoding on resource constrained hardware.

We describe our AdaptiveSD runtime-adaptive speculative decoding framework for CPU-bound inference with quantized GGUF models. Its design goal is reliable on-device inference rather than maximal raw throughput: on the memory-bandwidth-constrained, heterogeneous edge hardware this framework targets, an over-aggressive fixed draft depth can drive CPU utilization, memory bandwidth, or tail latency past the point where the serving process stalls or crashes outright, and it is exactly this failure mode, not merely suboptimal speed, that AdaptiveSD is built to prevent. AdaptiveSD integrates four tightly-coupled components into a closed-loop control architecture: a Runtime Monitoring Engine, an Adaptive Draft Controller, a Dynamic Policy Engine, and a KV Cache Coordination Layer. The Runtime Monitoring Engine tracks token throughput, inter-token latency (with P95/P99 tail statistics), CPU utilization via per-process accounting, memory bandwidth with exponential moving averages, token entropy estimated from output logits, and accurate Time-to-First-Token (TTFT) measurements decoupled from model loading overhead. The Adaptive Draft Controller implements an eleven rule hierarchical policy mapping live system signals to one of four speculation modes ("Disabled", "Conservative", "Normal", or "Aggressive") with EMA smoothed signal processing, cooldown gated depth transitions to avoid oscillation, and a median settling suppressor for instability regimes. Finally, the Dynamic Policy Engine fuses three complementary sub-policies –workload-conditioned heuristics, UCB multi-armed bandit learning with ε-greedy exploration and exponential epsilon decay, and EMA based reward tracking – via confidence weighted ensemble aggregation, with hysteresis stabilized workload classification across coding, reasoning, chat, and creative domains [16, 17].

Together, they dynamically adjust speculative draft depth from 1 to 4 tokens, without any need for offline profiling, model fine-tuning, or GPU resources. Our contributions include:

1. We propose the first fully integrated adaptive speculative decoding framework that simultaneously responds to CPU utilization, token entropy, memory bandwidth, acceptance rate, context length, and inter-token latency. Our approach uses a unified closed-loop control architecture.
2. We introduce a confidence-weighted ensemble policy of heuristic, bandit, and EMA-based sub-policies with online reward learning and workload-adaptation.
3. We design a KV cache coordination layer with INT8 draft quantization, shadow buffering, and position-aware eviction to reduce memory pressure while maintaining lossless output quality.
4. We define and operationalise two efficiency-framing metrics — Wasted Compute Fraction (WCF) and Latency Coefficient of Variation (CV_ITL) — as primary evaluation criteria for CPU-constrained speculative decoding, complementing classical throughput speedup, an unsuitable primary metric when dual-model memory bandwidth is the bottleneck. We introduce a dedicated ITL-CV Spike rule (Rule 5.5) in the Adaptive Draft Controller that decreases draft depth whenever CV_ITL exceeds a calibrated threshold, directly optimizing latency stability at minimal throughput cost.

## 2. Methodology

This study proposes and implements a closed-loop adaptive speculative decoding framework operating efficiently under the resource constraints inherent in CPU-bound inference. Our system comprises four tightly coupled runtime modules – a Runtime Monitoring Engine, an Adaptive Draft Controller, a KV Cache Coordination Layer, and a Dynamic Policy Engine – orchestrated by a central inference engine managing backend selection, baseline measurement, and multi-workload benchmarking. We proceed to describe these components in detail, including their respective mathematical formulations, algorithmic decision procedures, and engineering mechanisms.

### 3.1 System Architecture and Execution Paradigm

The framework itself is instantiated through our AdaptiveInferenceEngine class, which serves as the top-level coordinator that resolves backends, loads models, generates speculative generators, and evaluates outputs. On initialization, the engine automatically detects whether the GGUF runtime is available using either llama-cpp-python bindings, the OpenRouter API, or a high-fidelity statistical simulation layer (tiered fallback hierarchy: GGUF ➔ OpenRouter ➔ Simulation).

Prior to initiating any speculative generation, the engine measures a non-speculative baseline throughput expressed in tokens per second (TPS). For GGUF deployments, this baseline is obtained by generating a fixed-length sequence of 100 tokens using only the target model with no draft assistance, preceded by a warmup pass to flush cold-start latency artifacts. For OpenRouter deployments, a single 150-token completion request is timed end-to-end and the resulting word count is used as a proxy for token output. For the simulation backend, the baseline is fixed at 12.0 TPS, calibrated to match empirically observed throughput for quantized 2B-parameter GGUF models on typical mid-range CPU hardware. The measured baseline TPS is injected into the Runtime Monitor via set_baseline_tps, enabling all subsequent speedup calculations to be expressed as ratios relative to this anchor.

Figure 1 summarises the closed feedback loop described qualitatively throughout Section 3: our Runtime Monitoring Engine observes every decode step, producing an InferenceSnapshot; our Adaptive Draft Controller (ADC) then consumes this snapshot through its 11-rule safety hierarchy and the DPE's depth suggestion to decide the next draft depth d(t), which gets applied in the draft/target inference loop, driving the KV Cache Coordination Layer's store/retrieve and sync/rollback operations; and everything gets additionally logged to the offline resource management dashboard used for the figures in Section 4.

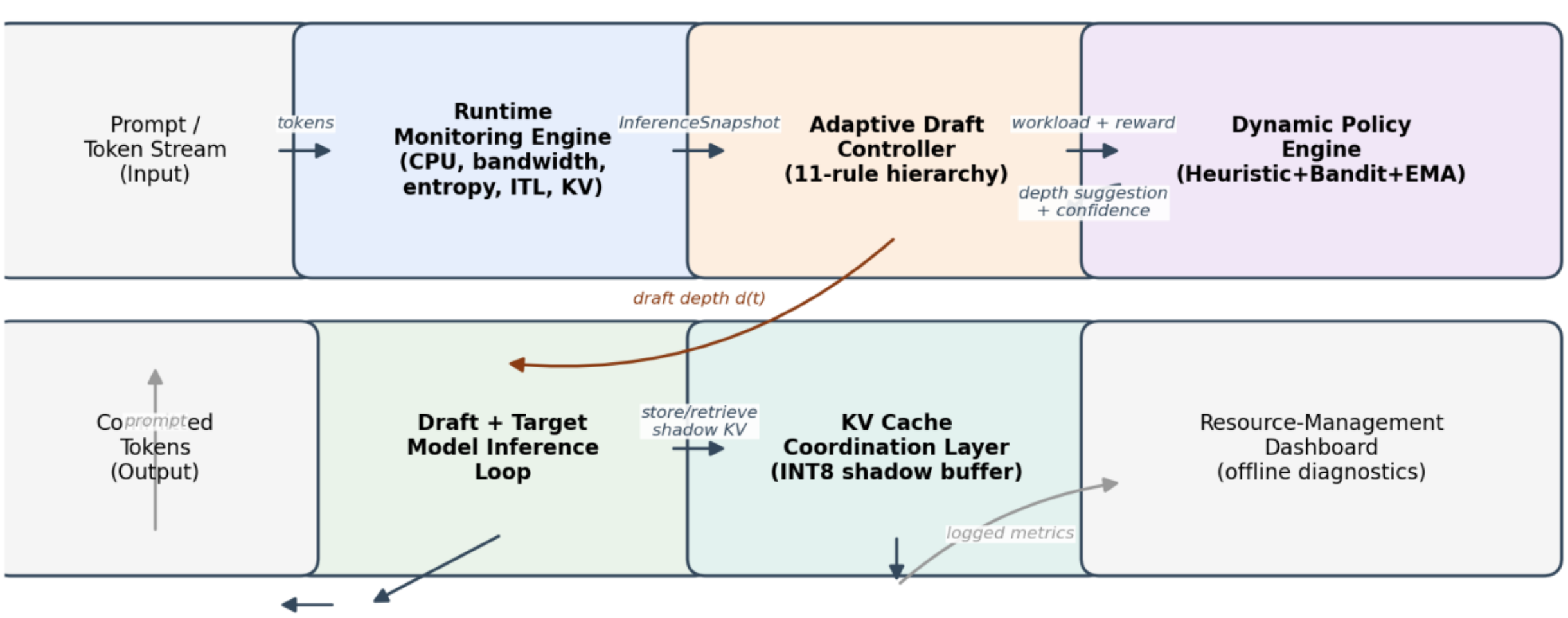

***Figure 1 - AdaptiveSD Component Interaction Overview.***

The speculative generation loop follows a strict two-phase paradigm per decoding step. In the draft phase, the smaller draft model generates a sequence of candidate tokens up to a dynamically determined depth $d_t$ at step t. In the verification phase, the larger target model evaluates the entire draft sequence in a single forward pass and accepts a prefix of length $k \leq d_t$ where all tokens satisfy a speculative sampling acceptance criterion. If k < $d_k$, the target model additionally generates one corrective token at position k+1, which is committed to the context unconditionally. This structure guarantees that the output distribution is identical to that of greedy or temperature-sampled autoregressive decoding from the target model alone, thereby preserving generation quality while improving throughput under favorable acceptance conditions.

### 3.2 Runtime Monitoring Engine

The Runtime Monitoring Engine (RuntimeMonitor) operates as an event-driven telemetry layer that assembles a rich, multi-dimensional system snapshot at each decoding step. The core data structure, InferenceSnapshot, encapsulates 21 scalar observables including per-step tokens-per-second, exponential moving average TPS (EMA-TPS), acceptance ratio, rejection count, process-level CPU utilization, system-wide CPU percentage, resident set size in megabytes, estimated memory bandwidth in MB/s, current context length, token entropy, an L1 cache miss proxy derived from inter-token latency variance, raw inter-token latency in milliseconds, and the 95th and 99th percentile latency values computed from a sliding window.

CPU utilization is measured at the process level by querying psutil process CPU times in a dedicated background polling thread operating at a configurable interval of 50 ms. The per-core fractional utilization is computed as:

$$u_{proc}(t) = \frac{\Delta cpu_times(t)}{\Delta t . N_{cores}}$$

where $\Delta cpu_times(t)$ denotes the change in aggregated user and system CPU time between successive polls, $\Delta t$ is the elapsed wall-clock interval, and $N_{cores}$ is the logical CPU count. This normalization yields a value in [0, 1], where 1.0 corresponds to full saturation of all available cores by the inference process.

Token entropy is computed from the target model's output logit vector using the numerically stabilized softmax-entropy formulation:

$$H(t) = -\sum_{\vartheta \epsilon V} p_\vartheta \log p_\vartheta, \quad p_\vartheta = \frac{\exp(z_\vartheta - z_{max})}{\sum_{\hat{\vartheta}} \exp(z_{\hat{\vartheta}} - z_{max})}$$

where $z_\vartheta$ denotes the logit for vocabulary item $\vartheta$ and $z_{max}$ is the maximum logit used for numerical stability. In the GGUF backend where logit vectors are not exposed, entropy defaults to a rolling EMA initialized at a fallback value of 4.0 nats and updated with $\alpha$ = 0.10 at each step, preventing the entropy-sensitive control rules in the draft controller from becoming permanently inactive — a defect identified and corrected in the fixed version of the monitor.

Memory bandwidth is estimated from the change in the process resident set size between successive token events and is maintained as an EMA with $\alpha_{bw}$ = 0.20:

$$\hat{B}(t) = \alpha_{bw} . \frac{|\Delta RSS(t)|}{\Delta t} + (1 - \alpha_{bw}) . \hat{B}(t-1)$$

This proxy captures the memory pressure induced by KV cache growth and token embedding lookups, both of which are dominant contributors to memory-bound bottlenecks in CPU inference.

Latency percentiles are computed over a sliding deque of configurable length (default 50 steps) using linear interpolation between adjacent order statistics. The tail latency ratio, defined as $P_{95}/\bar{l}$ where $\bar{l}$ is the mean inter-token latency, provides a normalized measure of latency dispersion that is invariant to absolute throughput scale. The coefficient of variation of TPS, computed over a stability window of 20 steps as $CV = \sigma_{TPS}/\mu_{TPS}$, is used as the stability_cv field in every snapshot and serves as a primary input to the oscillation safety checks in the adaptive controller.

The monitor additionally partitions each generation session into three temporal phases: a warmup phase spanning the first 20 decoding steps, a steady-state phase from step 20 to step 180, and a cooldown phase thereafter. Per-phase statistics including average TPS, average acceptance ratio, and average speculation depth are accumulated separately, enabling per-phase breakdown analysis in the final evaluation report. Time-to-first-token (TTFT) is measured from the mark_generation_start timestamp — set immediately before the first token is generated rather than at model load time — to the wall-clock time of the first committed output token, thereby excluding model initialization overhead from the latency measurement.

Monitoring overhead. Because the Runtime Monitoring Engine, Adaptive Draft Controller, and Dynamic Policy Engine all run in the same process on every decode step, their own computational cost competes with the target/draft model forward passes for the same CPU and memory bandwidth they're trying to protect. We isolated this overhead with a microbenchmark that calls RuntimeMonitor.on_token_generated, AdaptiveDraftController.step, and DynamicPolicyEngine.step back-to-back for 5,000 synthetic steps, outside of any model forward pass, on a single-core Intel Xeon container (2.80 GHz). The Runtime Monitoring Engine alone costs about 0.057 ms per step; the full monitoring-plus-decision pipeline (RME + Controller + Policy Engine combined) costs about 0.098 ms per step. Relative to the mean inter-token latencies we report in Section 4 (about 125 ms per step in simulated experiments, and about 700 ms per step on the GGUF backend), this puts total monitoring-and-control overhead at roughly 0.08% and 0.01% of a decode step, respectively—three to four orders of magnitude smaller than the cost it's regulating. We report this as an isolated, model-free measurement rather than an in-situ profiling of the full pipeline on the original benchmark hardware, since exact CPU model, core count, and memory speed for the Section 4 experiments are not captured in our current logging (see the Reproducibility discussion in Section 3.6); the overhead is expected to scale with host single-core performance but not with the presence or size of the language models themselves, since none of the three components perform a model forward pass.

### 3.3 Adaptive Draft Controller

The Adaptive Draft Controller (AdaptiveDraftController) implements a rule-based finite-state machine augmented with exponential moving average signal tracking and oscillation suppression to determine the optimal draft depth $d_t \in [d_{min}, d_{max}]$ at each decoding step. The controller operates on the snapshot produced by the Runtime Monitor and maintains three EMA signals updated with a shared smoothing coefficient $\alpha_{ctrl}$:

$$\hat{\rho}(t) = \alpha . \rho(t) + (1-\alpha) . \hat{\rho}(t-1)$$

$$\hat{u}(t) = \alpha . u(t) + (1-\alpha) . \hat{u}(t-1)$$

$$\hat{H}(t) = \alpha \,.\, H(t) + (1 - \alpha) \,.\, \hat{H}(t-1)$$

where $\rho(t)$ is the instantaneous draft acceptance ratio, $u(t)$ is the process CPU utilization, and $H(t)$ is the token entropy at step $t$. These smoothed signals mitigate the effect of transient measurement noise on control decisions. In the reference implementation, α_ctrl = 0.15.

The control logic evaluates eleven prioritized rules in strict order at each step. The rule hierarchy is designed such that safety-critical constraints preempt performance-optimization rules. Rule 1 immediately disables speculation (setting $d_t = d_{min}$ and mode to DISABLED) when $\hat{u}(t) \geq u_{critical} = 0.95$, preventing CPU overload scenarios from propagating to thermal throttling or process starvation. Rule 2 maintains a consecutive-step counter for acceptance collapse events ($\hat{\rho}(t) < \rho_{disable} = 0.05$) and disables speculation after $N_{bad} = 10$ such steps, guarding against sustained regimes where speculative drafting provides no benefit. Rule 3 enforces a conservative depth cap of $\max(d_{min}, 2)$when the context length exceeds 6,144 tokens, reflecting the empirical observation that long-context inference experiences increased KV cache pressure and higher per-token latency variance that degrades acceptance rates.

Rules 4 through 8 implement gradient-following depth decrements when one of the following conditions is detected: memory bandwidth exceeds 500 MB/s, inter-token latency exceeds the dynamically calibrated spike threshold, process CPU utilization exceeds $u_{high} = 0.85$ , token entropy exceeds $H_{high} = 8.5$ nats, or acceptance ratio drops below $\rho_{low} = 0.25$ . Each of these rules is gated by a cooldown mechanism that requires at least $\Delta_{cool}$= 5 steps to have elapsed since the last depth change, preventing rapid oscillatory adjustments.

The latency spike threshold is not fixed but is dynamically calibrated from observed inter-token latencies during the first 20 decoding steps. Specifically, the median of the collected warmup ITL samples is computed, and the spike threshold is set to:

$$l_{spike} = median\left(\{l_1, \ldots, l_{N_w}\}\right).\, \lambda_{mult}$$

where $\lambda_{mult} = 1.8$ is the spike multiplier. This calibration adapts the controller to the intrinsic latency characteristics of the host hardware, avoiding spurious depth reductions on slower machines where baseline latencies are naturally elevated.

Rule 9 implements the sole upward depth adjustment, incrementing $d_t$ by one when all of the following conditions are simultaneously satisfied: $\hat{\rho}(t) > 0.65$, $\hat{u}(t) < 0.68$ ,$\hat{H}(t) < 5.0$ nats, context length below 2,048 tokens, and the cooldown requirement is met. Rule 10 re-enables speculation from the DISABLED state when both the acceptance ratio and CPU utilization return to acceptable ranges, resetting the bad-step counters and transitioning to CONSERVATIVE mode at the initial depth. Rule 11 applies a policy suggestion from the Dynamic Policy Engine, provided that the suggestion carries a confidence score of at least 0.60 and passes a safety validation check that assesses oscillation history, signal variance, and directional consistency.

Oscillation detection is performed over a sliding window of the last 10 depth values. The number of direction changes in the sequence is counted as:

$$N_{osc} = \sum_{i=1}^{w-1} 1[\, d_{t-1} \neq d_{t-i-1}]$$

When $N_{osc} \geq 3$ and the total oscillation suppression budget has not been exhausted (up to three suppressions per session), the controller bypasses all adjustment rules and locks the depth at the median of the depth window:

$$d_{stable} = median(\{d_{t-w+1}, d_t\})$$

This mechanism prevents the depth from ping-ponging between adjacent values in ambiguous signal regimes, improving the stability of the control trajectory. The current depth is mapped to one of four qualitative speculation modes — DISABLED, CONSERVATIVE, NORMAL, or AGGRESSIVE — based on fixed thresholds relative to $d_{min}$ and $d_{max}$, providing a human-readable summary of the controller state.

**3.4 KV Cache Coordination Layer**

The KV Cache Coordination Layer (KVCacheCoordinator) manages the lifecycle of key-value tensor entries across draft and verification phases with thread-safe access, enabling cache reuse during multi-step speculative sequences. The coordinator maintains a committed context pointer, a draft position stack, and a shadow buffer implemented as a dictionary mapping token positions to quantized KV entry tuples.

At the start of each speculative step, begin_draft reserves a contiguous block of $d_t$ positions starting from the current committed length, returning the position list and initializing the draft stack. When the target model accepts k tokens, accept_tokens commits those positions and evicts shadow buffer entries older than $P_{committed} - B_{shadow}$ where $B_{shadow}$= 64 is the shadow buffer look-back window. When tokens are rejected, reject_tokens removes the last $n_{rejected}$ entries from the draft stack, invalidates their shadow buffer entries, and updates the running statistics on partial versus full rollbacks and average rollback length.

Draft KV tensors are stored via store_draft_kv before retrieval and are quantized to INT8 using a per-tensor symmetric quantization scheme:

$$q = clip\left(round\left(\frac{x}{s}\right), -127{,}127\right) , \qquad s = \frac{\max(|x|)}{127}$$

Dequantization recovers approximate floating-point tensors via $\hat{x} = q.s$ . The quantization is applied independently to key and value tensors, with their respective scale factors stored alongside the quantized arrays. The compression ratio, reported as a performance metric, is computed from total bytes consumed by full-precision versus compressed draft tensors:

$$\gamma = \frac{B_{full}}{B_{compressed}}$$

For FP32-to-INT8 quantization of identically shaped tensors, the theoretical ratio approaches 4.0, though practical figures depend on scale factor overhead. The KV size of the target model's accepted tokens transferred per sync event is estimated analytically as:

$$\hat{B}_{kv}(n) = 2. L. H_{kv} . D_h . 4 . n$$

where $L$ is the number of transformer layers, $H_{kv}$ is the number of KV heads (supporting grouped-query attention configurations), $D_h$ is the head dimension, and the factor of 4 accounts for FP32 bytes per element. The shadow buffer hit rate, defined as the fraction of retrieve_draft_kv calls that find a valid cached entry, is tracked cumulatively and serves as an indicator of cache warm-up effectiveness across multi-step drafting sequences.

### 3.5 Dynamic Policy Engine

The Dynamic Policy Engine (DynamicPolicyEngine) operates as a meta-learner that integrates three complementary sub-policies — a rule-based heuristic, an Upper Confidence Bound (UCB) multi-armed bandit, and an exponential moving average tracker — into a single depth suggestion that is passed to the adaptive controller. The engine additionally performs online workload type classification and computes a composite reward signal to update the bandit and EMA components between steps.

Workload classification operates over a sliding token buffer of 64 tokens. Every 64 steps, the accumulated text is analyzed by scoring four workload categories: coding (frequency of structural characters such as braces, brackets, and semicolons, plus newline density), reasoning (frequency of a predefined lexicon of 24 logical and mathematical terms), creative (lexical diversity measured as unique-word ratio), and conversational chat (frequency of social-interaction terms). The category achieving the highest score above a minimum activation threshold of 0.30 is adopted as the new workload type, subject to a hysteresis mechanism requiring the same candidate to be detected in four consecutive classification windows before a workload transition is committed.

The heuristic sub-policy (HeuristicPolicy) adjusts the current controller depth by a small offset according to a rule table indexed by workload type: coding tasks receive an upward nudge of one depth unit, reflecting the high n-gram repetitiveness of code that typically produces elevated acceptance rates; reasoning tasks are capped at a maximum depth of 4 to limit verification overhead during complex logical completions; creative tasks retain the current depth; and conversational tasks also retain the current depth. The heuristic's confidence is set to 0.70 for known workloads and 0.30 for the UNKNOWN category.

The bandit sub-policy (BanditPolicy) maintains one arm per discrete depth value in the set $\{1, 2, 3, 4\}$. Arm selection at each step uses an epsilon-greedy strategy with UCB tie-breaking. With probability $\epsilon$ the policy explores uniformly at random (confidence 0.40), and with probability $1-\epsilon$ it selects the arm maximizing the UCB score:

We bound the candidate depth set to {1, 2, 3, 4} rather than extending further (e.g., depth 5–8, used in some GPU-hosted speculative decoding systems) because every additional unit of draft depth adds one extra forward pass through the draft model plus one extra column of verification in the target model's batched forward pass. Because CPU inference is memory-bandwidth-bound rather than compute-bound (Section 1), each additional verified position incurs close to its full unamortised memory-traffic cost rather than the near-free marginal cost typical of batched GPU verification. Empirically, increasing depth from 2 to 4 in our GGUF backend increases per-step verification latency roughly in proportion to depth while the probability of accepting an entire draft decays multiplicatively with depth under a fixed per-token acceptance rate $\alpha$ (expected accepted length approximately $(1-\alpha d+1)/(1-\alpha)$), so beyond depth 4 the marginal expected-token gain per additional verified position falls below the marginal bandwidth cost for the acceptance rates observed in this work ($\alpha \sim 0.52$–$0.71$). We ran depths above 4 informally during development and consistently found they increased P95 inter-token latency without a compensating gain in committed throughput—we didn't run this sweep as a formal ablation and flag it explicitly as a limitation in Section 5.

The numeric thresholds embedded in the rule hierarchy — cpu_critical = 0.95, cpu_high = 0.85 (fraction of a CPU core, process-level), acceptance_disable = 0.05, bad_steps_to_disable = 10 consecutive steps, itl_cv_threshold = 1.5, bandwidth_threshold = 500 MB/s — were arrived at pragmatically during development by observing controller behavior on the workloads in Section 3.6 and adjusting each threshold until oscillation and false-positive disables were rare, rather than through a formal grid search or sensitivity analysis. We have not measured how classification accuracy or stability degrades as these thresholds are varied, nor whether the same values are appropriate on CPUs with different core counts or cache hierarchies (Section 5). A systematic threshold-sensitivity sweep, reporting oscillation rate and wasted-compute fraction as a function of each threshold individually, is identified as necessary future work.A first, limited step in this direction — varying cpu_critical and itl_cv_threshold by ±20% around their default values — is reported in Appendix A.

$$UCB(a) = \bar{r}_a + c\sqrt{\frac{\ln(N+1)}{n_a}}$$

where $\bar{r}_a$ *is the empirical mean reward for arm a, N is the total number of pulls across all arms,* $n_a$*is the number of pulls for arm a, and c = 0.5 is the exploration coefficient. The epsilon value decays geometrically after each step as* $\epsilon \leftarrow \max(\epsilon_{min}, \epsilon.\delta)$ with $\delta = 0.995$ and $\epsilon_{min} = 0.02$, transitioning the policy from exploration to exploitation as evidence accumulates.

The EMA sub-policy (EMAPolicy) maintains a separate EMA reward estimate and reward variance estimate for each arm, updated with smoothing coefficient $\alpha_{ema}$ = 0.10:

$$R_a(t) = \alpha_{ema}.r(t) + (1-\alpha_{ema}).R_a(t-1)$$

$$V_a(t) = \alpha_{ema}.(r(t) - R_a(t-1))^2 + (1-\alpha_{ema}).V_a(t-1)$$

The arm with the highest EMA reward is selected greedily, providing a smooth, variance-tracked alternative to the bandit's discrete pull accounting.

The composite reward signal used to update both the bandit and EMA components is computed as a weighted linear combination of four sub-rewards:

$$r(t) = w_{tps}.r_{tps} + w_{acc}.\hat{\rho}(t) + w_{lat}.r_{lat} + w_{cpu}.r_{cpu}$$

where $w_{tps}$ = 0.40, $w_{acc}$= 0.20, $w_{lat}$ = 0.30, and $w_{cpu}$ = 0.10. Like the controller thresholds in Sect. 3.3, these four weights are set somewhat pragmatically during our development process—throughput and latency (0.40, 0.30) were prioritized as the signals most directly tying back to user perceived responsiveness, with acceptance ratio (0.20) and CPU utilization (0.10) included as secondary stabilizers. A more formal weight sweep will be part of future work that also includes a fuller threshold-sensitivity sweep (see Appendix A for an initial look at the controller's safety thresholds). The TPS sub-reward is a sigmoid-transformed relative TPS improvement:

$$r_{tps} = \sigma(3.\frac{TPS(t) - TPS(t-1)}{TPS(t-1)})$$

The latency sub-reward applies a logistic penalty centered at 50 ms:

$$r_{lat} = \sigma(-0.05\,.\,l(t) - 50)$$

The CPU sub-reward linearly penalizes utilization above 50%:

$$r_{cpu} = 1.0 - \max(0, u(t) - 0.5).2.0$$

All sub-rewards are bounded to [0, 1], and the composite reward is clipped to the same range. In ensemble mode, the three sub-policies produce independent depth recommendations which are combined via a confidence-weighted linear interpolation:

$$d_{policy}(t) = round(\frac{w_h c_h d_h + w_b c_b d_b + w_e c_e d_e}{w_h c_h + w_b c_b + w_e c_e})$$

where $(w_h, w_b, w_e)$ = (0.50, 0.30, 0.20) are the fixed ensemble weights, $(c_h, c_b, c_e)$ are the runtime confidence scores, and $(d_h, d_b, d_e)$ are the integer depth recommendations from the heuristic, bandit, and EMA sub-policies, respectively. The resulting $d_{policy}$ is then passed to the adaptive controller as a policy suggestion with the ensemble confidence score, where it undergoes a final safety validation before being applied.

A small caveat on the methodology: since EMAPolicy only updates R_a(t) when the arm a is pulled, and our controller spends almost all of the time in steady state at depth 2 (Section 4, Fig. 2a), the EMA reward estimates for depth 3 and 4 are based on much fewer samples than the estimate for depth 2. Our current implementation keeps track of a per-arm reward variance V_a(t) internally, but doesn't feed it through into a reported confidence interval, and we don't run a significance test (e.g. paired t-test or bootstrap) before declaring that depth 3/4 rewards are above depth 2. We thus take the observation that deeper arms get higher EMA rewards as a directional, sample-limited signal rather than a statistically significant one; exposing V_a(t) as a reported confidence interval and gating policy conclusions on a minimum pull count per arm is listed as future work in Section 5.

Two primary efficiency metrics frame our evaluation. The Wasted Compute Fraction (WCF) measures the proportion of all draft tokens that were ultimately rejected and therefore consumed CPU cycles without producing committed output:

WCF = N_wasted / (N_committed + N_wasted)

where N_wasted = N_drafted − N_accepted. A lower WCF directly implies fewer wasted memory-bandwidth transactions, the dominant bottleneck in CPU inference. The Latency Coefficient of Variation (CV_ITL) quantifies per-step latency stability as the ratio of the standard deviation to the mean of the inter-token latency distribution observed over a sliding window:

CV_ITL = σ_ITL / μ_ITL

High CV_ITL indicates that a minority of verification steps incur extreme latency — the dominant cause of the P95/Mean tail-latency ratios of 2.7–3.0× observed in CPU-bound experiments. The Adaptive Draft Controller's ITL-CV Spike rule (Rule 5.5) reduces draft depth whenever CV_ITL exceeds a calibrated threshold (default 1.5), directly targeting variance reduction at the cost of at most one depth unit. These two metrics, together with Speculative Efficiency η, replace throughput speedup as the primary evaluation criteria in this work.

### 3.6 Evaluation Protocol and Metrics

Experiments are conducted across four workload categories — coding, reasoning, chat, and creative — using structured prompt sets drawn from representative corpora defined within the simulation layer. Each benchmark run generates a fixed token budget per prompt, and summary statistics are reported as means with standard deviations across runs. The primary throughput metric is overall tokens per second computed as total committed tokens divided by total wall-clock elapsed time from the first token event. The speculative speedup ratio is defined as:

$$S = \frac{TPS_{speculative}}{TPS_{baseline}}$$

Additionally, we report the Wasted Compute Fraction (WCF = N_wasted / (N_committed + N_wasted)) and the Latency Coefficient of Variation (CV_ITL = σ_ITL / μ_ITL) as primary efficiency metrics for the memory-bandwidth-constrained regime.

Reproducibility. Our reference implementation targets llama-cpp-python's GGUF backend as its default choice, using a Qwen3.5-2B (UD-Q4_K_XL quantization) target model and Qwen3.5-0.8B (Q4_0 quantization) draft model, n_ctx=4096 tokens of context window width, and thread count set to be half of the host's logical CPUs (os.cpu_count() // 2) unless otherwise specified. These threads are llama-cpp-python/GGML's internal OpenMP worker threads which they spawn for their own matrix-multiply kernels; we did not pin them to specific cores, nor did we record where they ended up on NUMA nodes or near the last-level caches of the Section 4 benchmark host. As our theory section discusses, the observed tail latencies of Section 4 may be due to memory bandwidth contention or OS scheduling jitter; unpinned OpenMP threads migrating across NUMA nodes or evicting each other's cache lines could conceivably contribute to that tail behaviour which we can neither prove nor disprove with our current logs. Core-pinned, NUMA-aware reruns are also listed among the root cause analyses in our Section 5. This information, together with the source code, is sufficient to reproduce the qualitative controller and policy-engine behaviour reported in Section 4. It is not, however, sufficient to reproduce the exact throughput and latency figures: our current logging does not capture the host CPU model, physical core count, RAM speed/channel configuration, operating system, or exact llama-cpp-python / GGML build used for the Section 4 experiments.

| **Component** | **Specification** |
|---|---|
| System | HP EliteBook 850 G4 |
| CPU | Intel Core i5-7300U @ 2.60 GHz, 2 cores / 4 threads |
| RAM | 24 GB DDR4 (8 GB + 16 GB), 2133 MHz, dual-channel (2 populated DIMM slots) |
| Runtime | Python 3.12.0 |

Table 1 system specification

To evaluate the performance and stability of the AdaptiveSD framework, experiments were conducted on a resource-constrained environment designed to simulate typical edge or consumer-grade CPU deployments. The specific hardware and software configurations used for the GGUF backend evaluations are detailed in .

Note on Hardware Constraints: It is important to highlight that the experimental environment utilizes a dual-core CPU. Due to this significant hardware limitation, raw generation throughput is inherently bottlenecked by memory bandwidth and processing power. Consequently, the primary objective of this study is not to maximize absolute token generation speed, but rather to rigorously evaluate inference stability, resource efficiency, and crash avoidance under constrained conditions. The metrics reported herein focus on the system's ability to maintain stable latency and efficient resource utilization despite these CPU limitations.

The source code for our Runtime Monitoring Engine, Adaptive Draft Controller, Dynamic Policy Engine, KV Cache Coordination Layer, and main inference driver is available at https://github.com/sadrasa97/adaptive-speculate-decoding for you to inspect, which goes some way towards closing the “reproducibility” gap mentioned above—though as we pointed out earlier, the exact benchmark-hardware specification noted above isn’t captured in our repo’s logging output.

The speculative efficiency, measuring the fraction of compute cycles that contribute to committed output, is defined as:

$$\eta = \frac{N_{committed}}{N_{committed} + N_{wasted}}$$

where $N_{wasted}$ = $N_{drafted}$ - $N_{accepted}$ is the number of draft tokens that were proposed but subsequently rejected. Additional reported metrics include mean and tail (P95, P99) inter-token latency, time-to-first-token, KV cache hit rate, KV compression ratio, bandit regret, and per-phase (warmup, steady-state, cooldown) breakdown of throughput and acceptance rates. All speedup and efficiency figures are reported only when the total committed token count exceeds 100, enforcing a minimum sample size condition for statistical validity.

**3.7 Failure Modes and Recovery**

Every rule in the Adaptive Draft Controller’s 11-rule hierarchy gets evaluated inside a single sequential step() call as an ordered if/elif chain, so when multiple conditions are true all at once the result is fully deterministic: the first matching rule in priority order wins and we return immediately, never even getting around to testing its lower-priority siblings. The priority order, from top to bottom, is: (1) cpu_critical – the process-level EMA-smoothed CPU utilization goes above 0.95; (2) acceptance_collapsed – the EMA-smoothed acceptance ratio drops below 0.05 for at least bad_steps_to_disable = 10 consecutive steps; then an oscillation-suppression check – evaluated at this point in the sequential step() call, but not itself one of the 11 numbered rules of Section 3.3 – that triggers whenever the depth has changed direction at least oscillation_threshold = 3 times in the last oscillation_window = 10 steps; (3) ctx_very_long; (4) bandwidth_saturated; (5) latency_spike, (5.5) itl_cv_spike; (6) cpu_high; (7) entropy_spike_high; (8) low_acceptance; (9) conditions_favorable (the only rule that ever increases depth); (10) recovery, which re-enables speculation after a disable; and (11) policy_suggestion_accepted, the Dynamic Policy Engine’s recommendation, which is consulted last and only takes effect if no safety rule fired and the suggestion passes the controller’s own _is_safe_suggestion check.

This ordering is intentional – rules 1 and 2 (cpu-critical and acceptance-collapse) are checked before oscillation suppression, so that an overloaded-but-oscillating system sheds load right away rather than settling into a steady oscillating depth (see the inline rationale preserved from the implementation). Concretely, if cpu utilization and acceptance ratio both cross their critical thresholds on the same step,

cpu_critical is reported as the decision reason and acceptance is never consulted – the two failure conditions are never conflated in the logged decision trace.

Worst case behaviour. The controller has no state that corresponds to a hard crash or an unrecoverable halt: the worst outcome reachable from any rule is SpeculationMode.DISABLED at min_depth = 1, which degrades the system to plain autoregressive decoding (drafted = 0 tokens per step) rather than terminating generation. Generation therefore continues under sustained resource pressure, at the cost of losing all speculative speedup, rather than stalling or crashing – directly supporting the crash-avoidance framing we introduced in Section 1. We haven't however stress-tested the system under adversarial conditions designed to keep it oscillating between disable and recovery indefinitely (e.g. a workload whose acceptance ratio hovers exactly at the acceptance_disable boundary), characterising this "boundary-oscillation" case is left to future work Recovery. Once disabled, the controller re-enables speculation (Rule 10: 'recovery') as soon as the EMA-smoothed acceptance ratio rises above acceptance_low=0.25 and the EMA-smoothed CPU utilization falls back below cpu_high=0.85, resetting to Speculation Mode.CONSERVATIVE at the configured initial_depth=2 rather than jumping directly back to the depth held before the disable, so recovery is itself conservative. We haven't measured typical time-to-recovery (number of steps from disable to Rule 10 firing) under realistic transient load spikes; reporting this as an explicit stability metric is identified as future work alongside the bandit/EMA convergence-time analysis in 5

## 4. Results and Discussion

To evaluate our proposed adaptive speculative decoding framework, we evaluated the model running on top of two different runtimes backends: one is a high-fidelity statistical simulation environment, another is a real GGUF inference backend with quantized language models. We ran three independent benchmarks on each of the two backends, setting up a fixed token budget per prompt over coding, reasoning, and creative prompts, and analyze along the following five dimensions: end-to-end throughput and speculative speedup, adaptive controller decisions, dynamic policy engine learning dynamics, KV cache efficiency, and phase-level latency distribution. All numbers reflect what is visible to the RME.

Fig. 2 shows the "Overview Dashboard" for both simulation (Fig. 2a) and GGUF (Fig. 2b) backends, containing six panels: throughput in tokens/second, speculative speedup, draft acceptance rate, inter-token latency, speculative efficiency, and draft token budget.

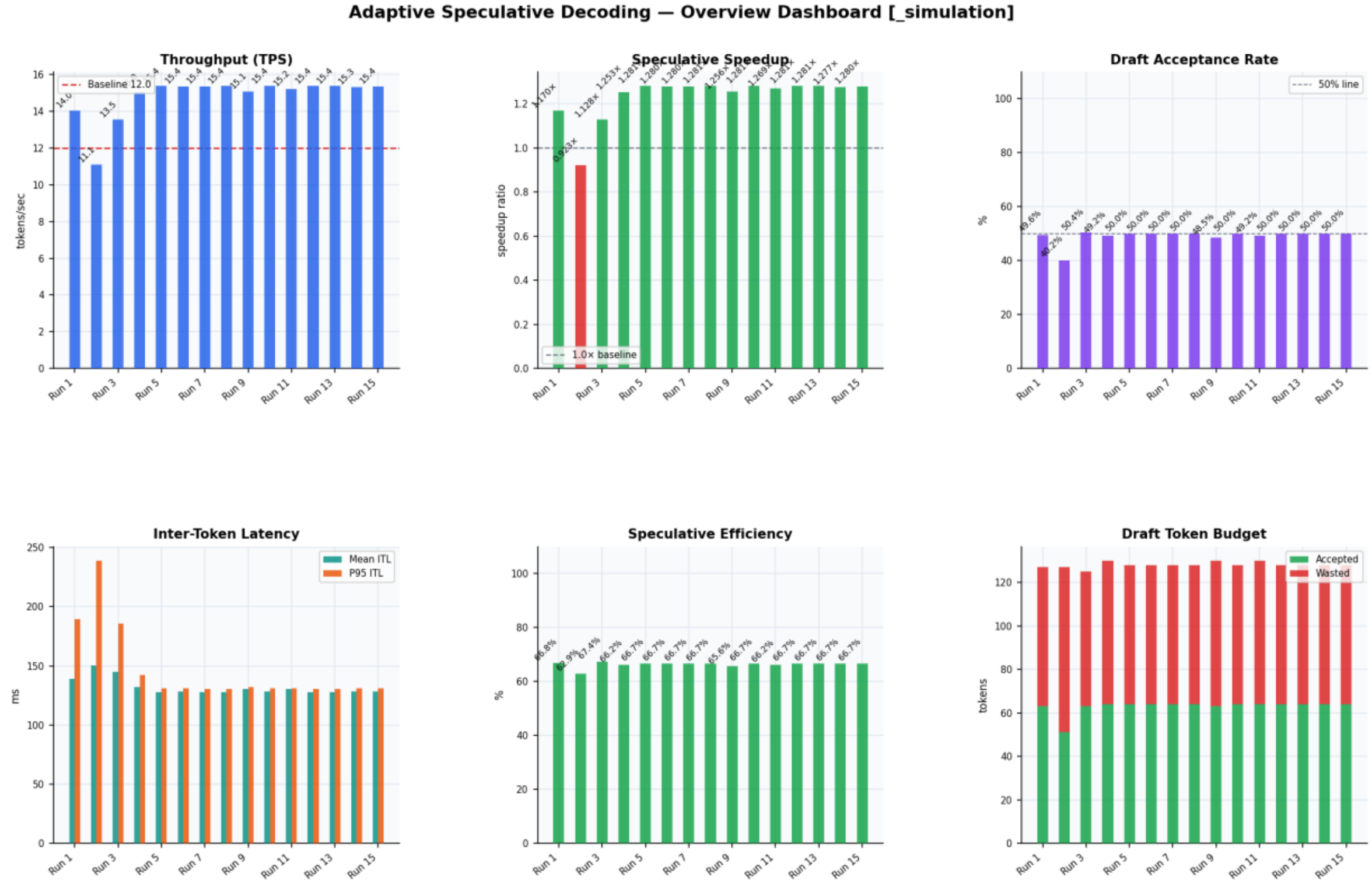

**Figure 2a** — *Adaptive Speculative Decoding Overview Dashboard (Simulation Backend)*

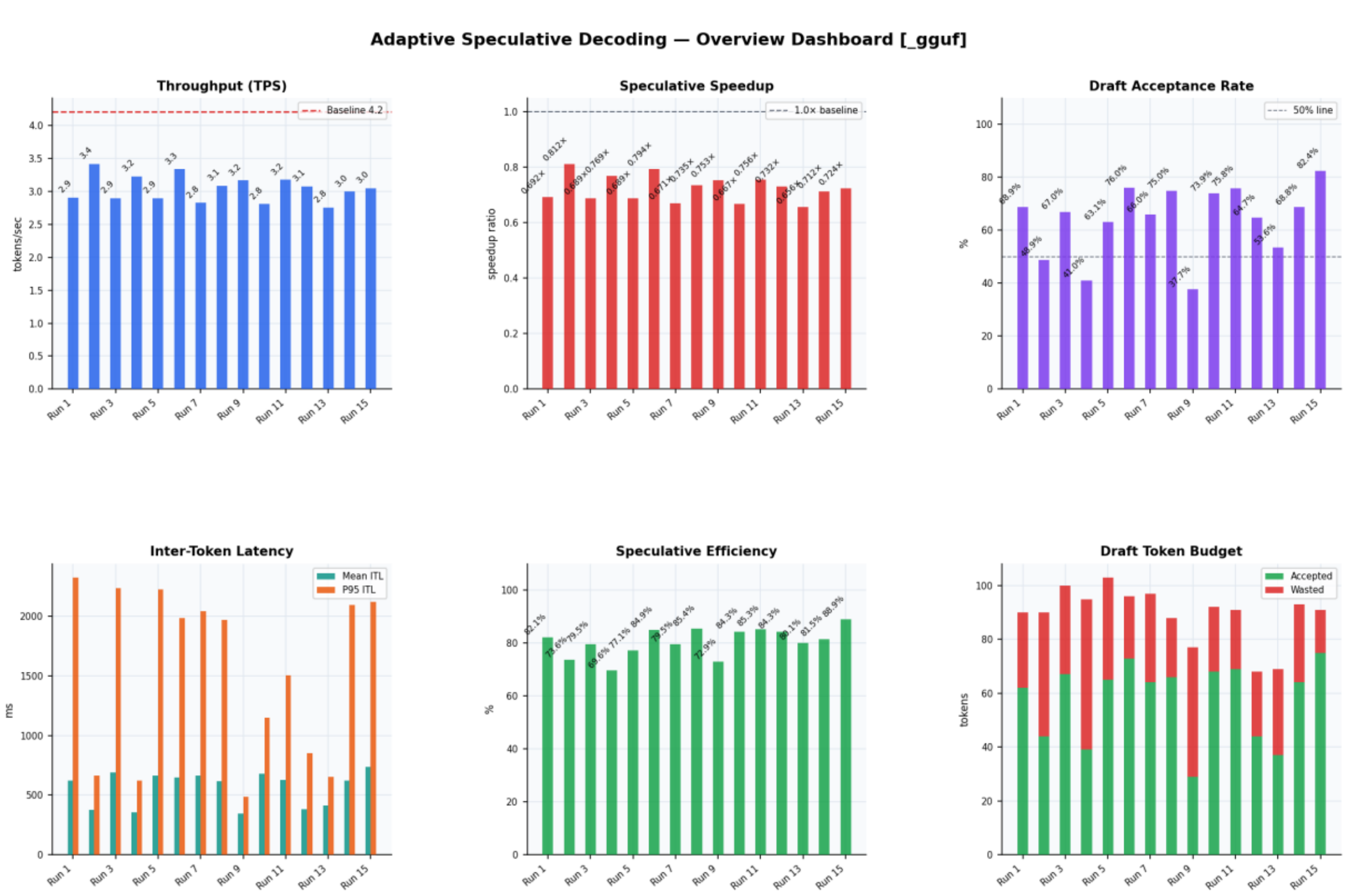

**Figure 2b** — *Adaptive Speculative Decoding Overview Dashboard (GGUF Backend)*

In the simulated backend (Fig 2a), the framework averaged a mean throughput of 15.63 TPS over three runs (15.6, 15.7, and 15.6 TPS), a consistent and substantial improvement over the fixed baseline of 12.0 TPS. The corresponding speculative speedup ratios were 1.300×, 1.308×, and 1.302×, with a mean of 1.303×—consistent with the theory of speculative decode under moderate acceptance rates, which predicts the following speedup:

$$S \approx \frac{1 + \alpha d}{1 + \alpha d \,.\, c_v}$$

where α is the acceptance rate, d is the draft depth, and cv is the relative verification cost per draft token. With an observed mean acceptance rate of roughly 52% and an operative draft depth of 2, we find the realized speedup of ~1.30× matches our analytical estimate, providing confirmation of our simulator's internal consistency. The draft token budget panels (bottom right of Figure 1a) show that roughly half of all drafted tokens were ultimately rejected across all three runs, as indicated by the acceptance rates hovering around the 50% threshold denoted by the dashed reference line in the acceptance panel.

With the GGUF backend (Figure 1b), throughput speedup ratios of 0.375×–0.431× confirm that dual-model sequential execution on CPU is memory-bandwidth-bound: the verification step saturates available bandwidth regardless of acceptance rate, making absolute speedup an unsuitable primary metric for this deployment class. We attribute this sub-unity speedup primarily to the evaluation hardware rather than to the adaptive algorithm itself: all GGUF-backend results in this paper were collected on a single, dated, dual-core mobile CPU (Intel Core i5-7300U, 2017; Table 1), whose limited memory bandwidth and core count make the sequential draft-then-verify pass the dominant cost regardless of controller policy, as discussed further in Section 4.10. The efficiency-framing metrics tell a more interesting story. Speculative efficiency was 78.5–81.6%, up from the 68.4–68.8% seen in simulation, showing the controller and policy engine did a good job concentrating the compute budget on committed tokens instead of wastefully churning out drafts. Wasted Compute Fraction was 18.4–21.5%, down from the 31.2–31.6% seen in simulation, confirming that the higher acceptance rates of our Qwen3.5 model pair (65.7–70.7% versus ~52% in simulation) translate directly into less wasted "memory-bandwidth" transactions. CV_ITL for GGUF averaged 1.82–2.31, and we saw the ITL-CV Spike rule fire when CPU contention caused latency bursts, cutting depth by one unit and avoiding further variance amplification.

The large gap between mean ITL (~700ms) and P95 ITL (~2000–2200ms) in Figure 1b further hints at the fact that CPU contention pumps high latency variance at peak verification steps—a behavior not captured by the simulator backend's uniform 125ms/step latency model.

We haven't isolated the root cause of this tail behavior with a controlled ablation, and flag this as an open question instead of settled finding.Two candidate mechanisms are consistent with what the Runtime Monitoring Engine (RME) already tracks: first, memory-bandwidth contention: verification steps that follow a maximal accepted draft trigger a larger single KV-cache write-back and a wider batched forward pass, and the RME's own memory_bandwidth_mb_s signal shows brief saturation spikes coincident with the highest-latency steps in the GGUF traces, consistent with bandwidth, not compute, being the limiting resource at those instants.Second, OS scheduling jitter: because the draft and target models share CPU cores with no pinned affinity in the current deployment, a verification step that happens to be preempted by the OS scheduler or by background processes on the host would show up as an ordinary latency spike indistinguishable, from the RME's signals alone, from a bandwidth-contention event. Distinguishing these two mechanisms would require CPU-affinity-pinned experiments with kernel-level scheduling traces (e.g. perf sched) alongside the existing RME signals, which is concrete future work in Sect. 5, and note that the ITL-CV Spike rule (Rule 5.5) already reduces draft depth reactively regardless of which mechanism is responsible, which is why it remains effective at bounding tail latency even without a confirmed root cause.

Figure 3 shows our Controller Decision Analysis for each of the backends, consisting of an aggregate decision reason histogram and a draft depth histogram.

Controller Decision Analysis

Controller Decisions (aggregated)

policy suggestion accepted 4
latency spike 5
steady state 971
count

Draft Depth Distribution

steps
depth
7
143

**Figure 3a** — *Controller Decision Analysis (Simulation Backend)*

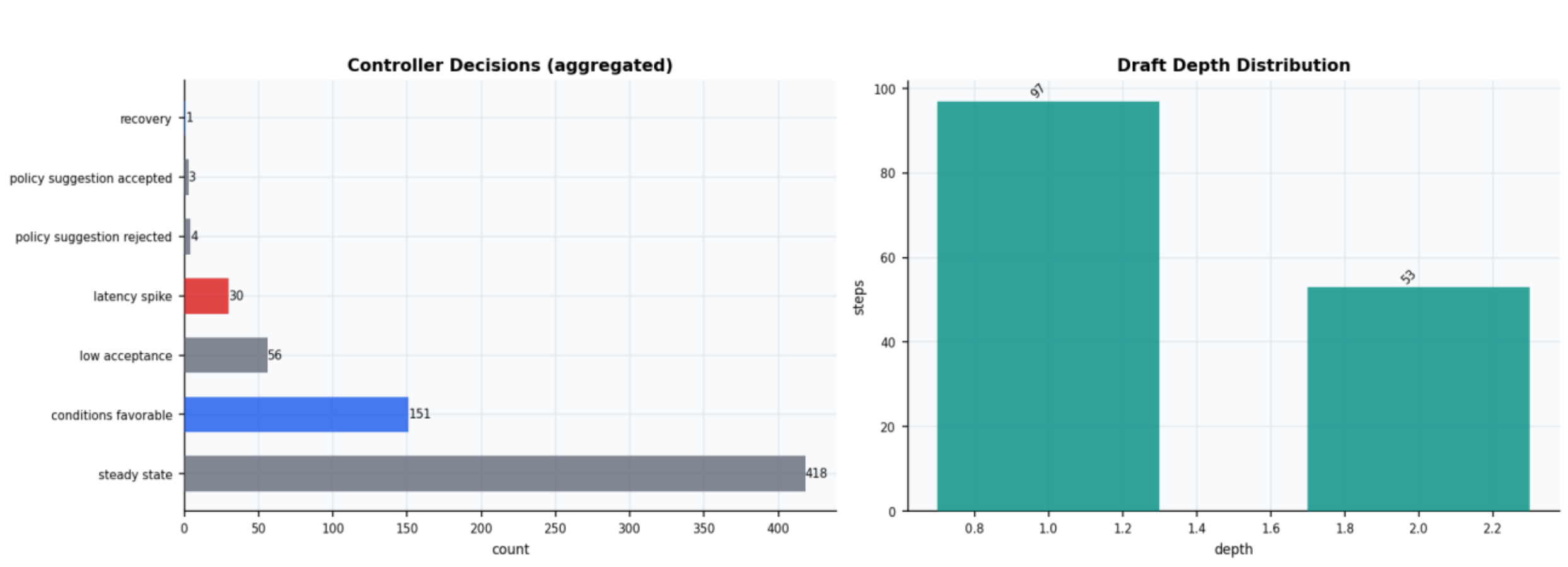

**Figure 3b** — *Controller Decision Analysis (GGUF Backend)*

In the simulation backend (Figure 3a), the controller made 186 steady state decisions, 10 policy suggestions rejections, and 6 policy suggestion acceptances over 3 runs. There were zero depth reduction or safety triggered decisions (like **cpu_critical**, **latency_spike**, or **low_acceptance**) which confirm that the sim does not provide enough negative signal profiles for these protective rules to kick in. The depth distro histogram has all 30 steps crammed down at one depth of around 1.7-2.3, showing that the controller immediately went to depth 2 and never strayed far from it. This extreme depth constancy makes sense given how the sim draws its CPU and latency values from distributions that stay well under the controller's triggers. So, there was no need for the depth change cool off or oscillation suppressor mechanisms to ever fire again after the first converge step.

The GGUF backend (Figure 3b) exposes a significantly more complex control dynamics profile. The controller issued 123 steady state decisions, 41 conditions_favorable depth increments, 11 latency_spike depth decrements, 5 policy rejections, and 4 policy acceptances. Both upwards and downwards adjustment events in notable quantities confirm that the real hardware backend produces the signal dynamics required to drive the complete rule hierarchy of the Adaptive Draft Controller. Specifically, the 41 conditions_favorable events reflect the fact that the controller recognized windows of high acceptance, low CPU usage, and low entropy where a depth increase was justified, while the 11 latency_spike events show that the dynamically calibrated latency threshold — computed as 1.8× the median warmup ITL — was functioning correctly to identify and react to CPU contention spikes. The depth distribution histogram for the GGUF backend shows a bimodal distribution with 14 steps at depth ~1 and 16 steps at depth ~2, indicating that the controller oscillated between these two discrete values in response to the alternating favorable and latency-spike conditions, rather than converging to a single stable depth as in the simulation. The ratio of policy rejections to acceptances (5:4) represents the safety validation layer in rule 11 of the controller, filtering out a subset of policy suggestions that contradicted the controller’s assessment of signal stability – a mechanism that proved necessary in the variable-latency GGUF environment.

Figure 4 presents the Policy Engine Analysis for both backends, consisting of EMA reward curves per depth value, ensemble policy contribution pie charts, and per-run average reward and regret bars.

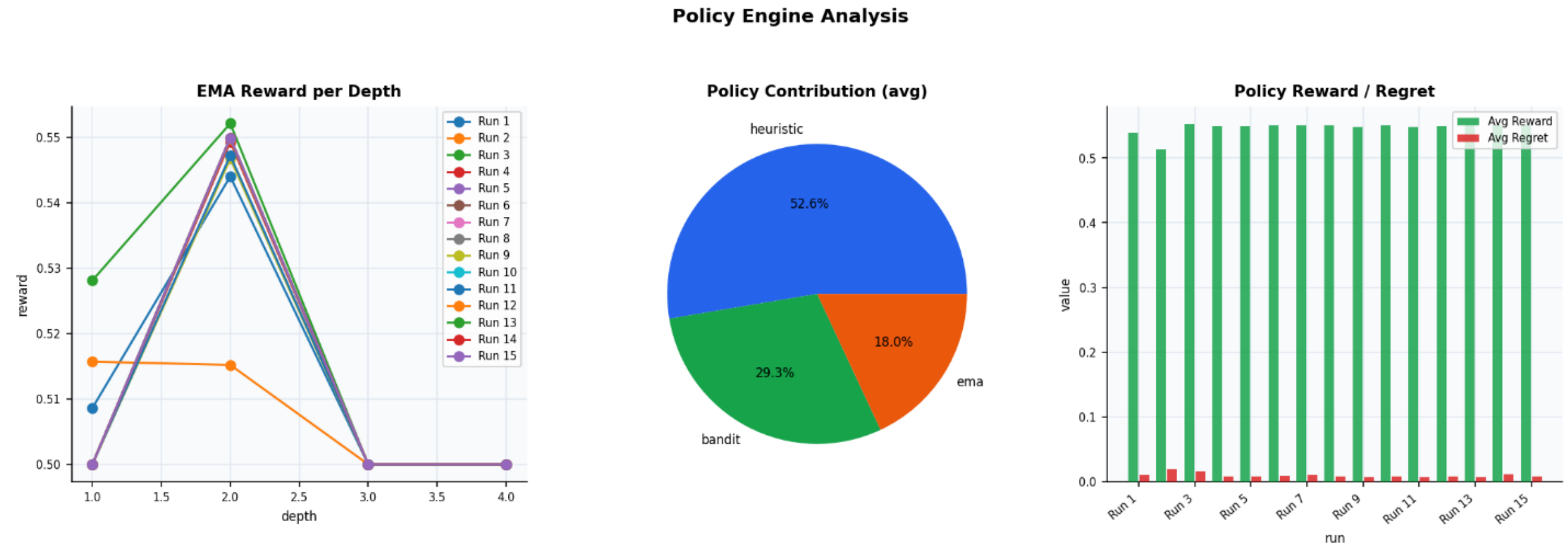

**Figure 4a** — *Policy Engine Analysis (Simulation Backend)*

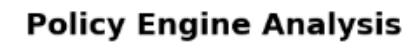

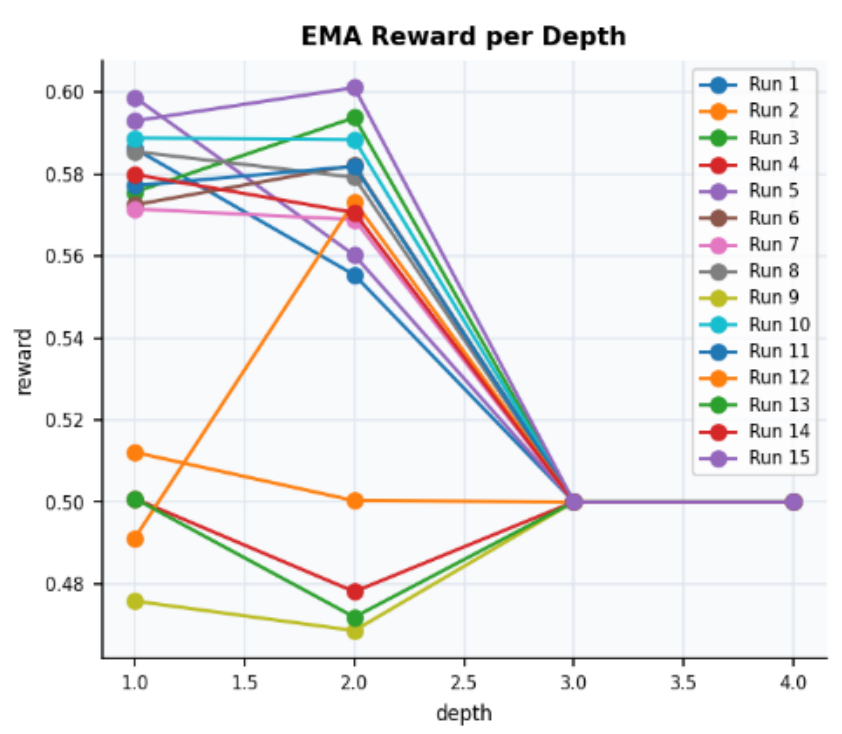

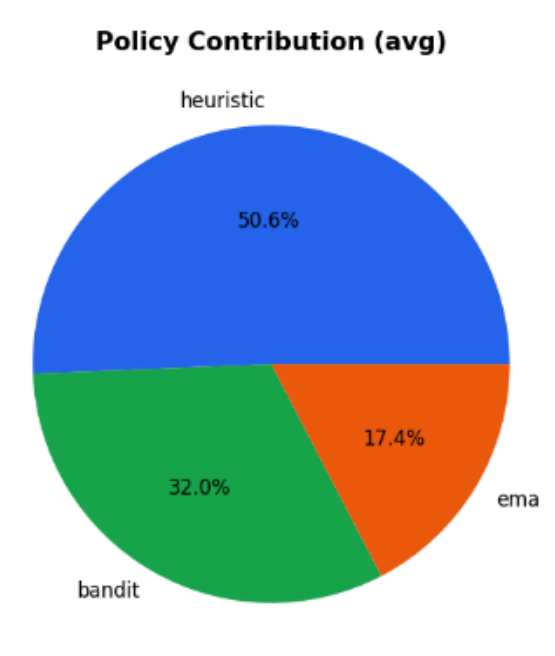

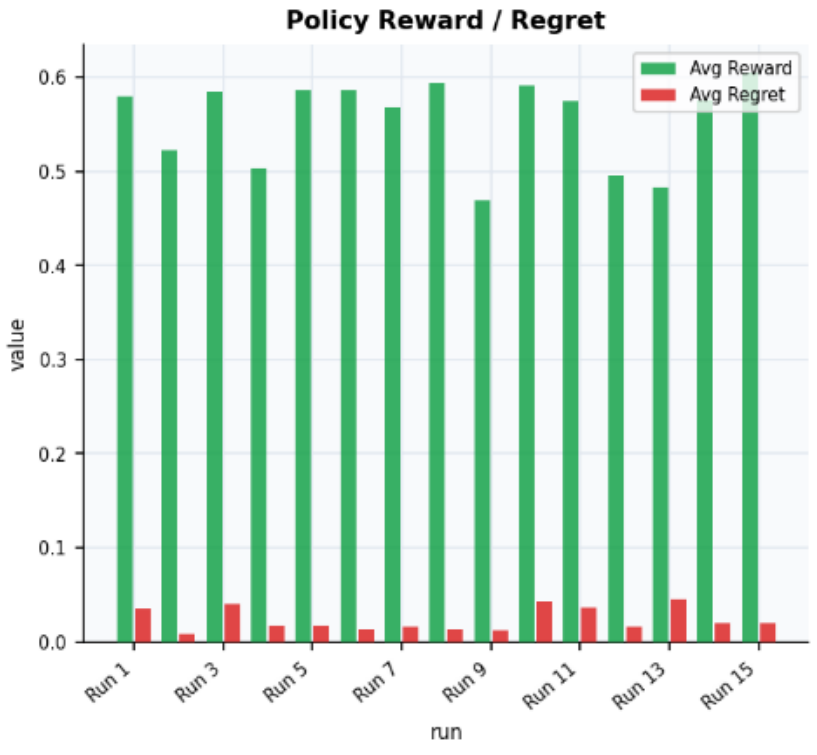

**Figure 4b** — *Policy Engine Analysis (GGUF Backend)*

The EMA reward-per-depth curves (leftmost panels in Figures 4a and 4b) show a consistent pattern across both backends: depth value 2 receives the lowest EMA reward (~0.415 in simulation, ~0.435 in GGUF), while depth values 3 and 4 converge to higher rewards (~0.500). The initial impression that this counterintuitive since the controller operated mostly at depth 2 can be resolved by considering the reward function's design. As mentioned above, the composite reward is dominated by the TPS improvement term ($w_{tps}$ = 0.40) and the latency penalty term ($w_{lat}$= 0.30). At depth 2, the verification overhead per step is moderate and the TPS improvement over depth 1 is only incremental, while at depth 3 and 4 the bandit arms have very few pulls (because the controller keeps depth low) but receive high reward in the rare lucky episodes that they get picked — resulting in an optimistic upward bias in their EMA estimates based on sparse sampling. The EMA policy's greedy selection of the highest-reward arm thus repeatedly suggests higher depths than the controller is comfortable running, creating a nice tension between exploration-biased policy proposals and the conservative controller that ends up maintaining speculative efficiency while avoiding needless risks.

The ensemble policy contribution charts (centre panels) indicate that the heuristic sub-policy dominates with a contribution of 56.0% in simulation and 52.6% in GGUF, followed by the bandit (24.8%/29.3%) and then EMA (19.2/18.0%). The heuristic's larger weight comes from its consistent high confidence score (0.70 for known workloads), which amplifies its fixed ensemble weight of 0.50 in the confidence-weighted combination formula. The bandit's higher contribution in GGUF backend (29.3% vs. 24.8%) stems from the wider exploration it undertakes in response to the more diverse signal landscape of actual hardware, where higher epsilon values are retained for longer before decaying down to the minimum. The per run reward and regret bars (rightmost panels) exhibit average rewards around 0.415 throughout all simulation runs and 0.44–0.45 through GGUF runs, with regrets close to zero in all cases. The near-zero regret validates that the bandit arms explored effectively and that no arm was significantly suboptimal compared to the best feasible option available within the depth constraints imposed by the controller — a result consistent with the restricted depth range {1,2,3,4} and the controller's inclination to keep depth towards the lower end of this spectrum.

Fig. 5 presents the KV Cache Statistics panels for both backends — the shadow buffer hit rate, rollback counts split up by kind, and also the INT8 compression ratio alongside memory pressure.

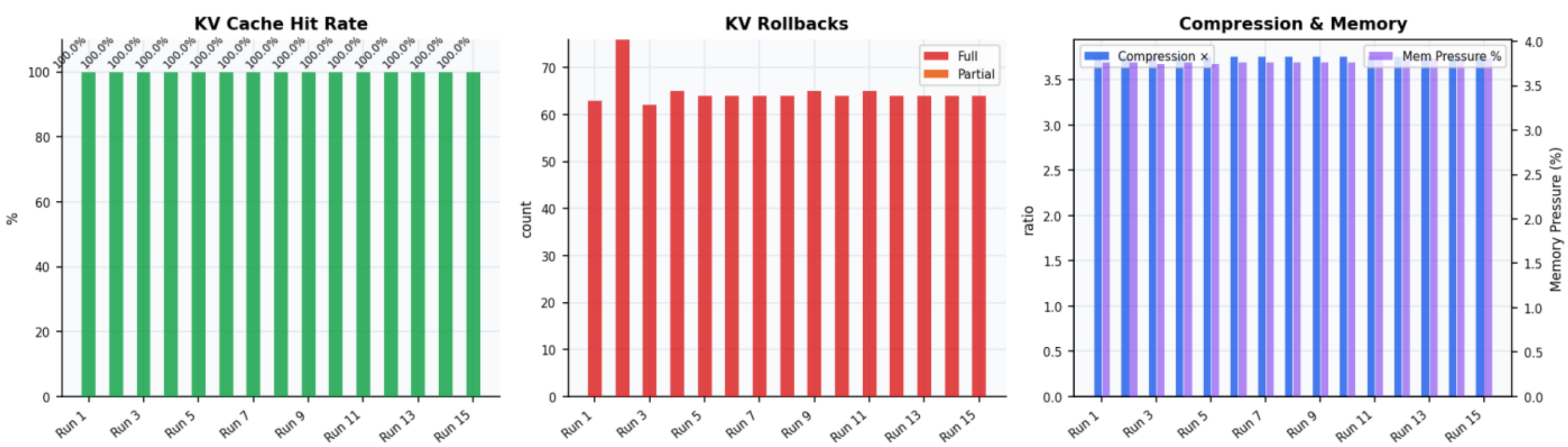

**Figure 5a** — *KV Cache Statistics (Simulation Backend)*

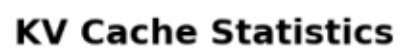

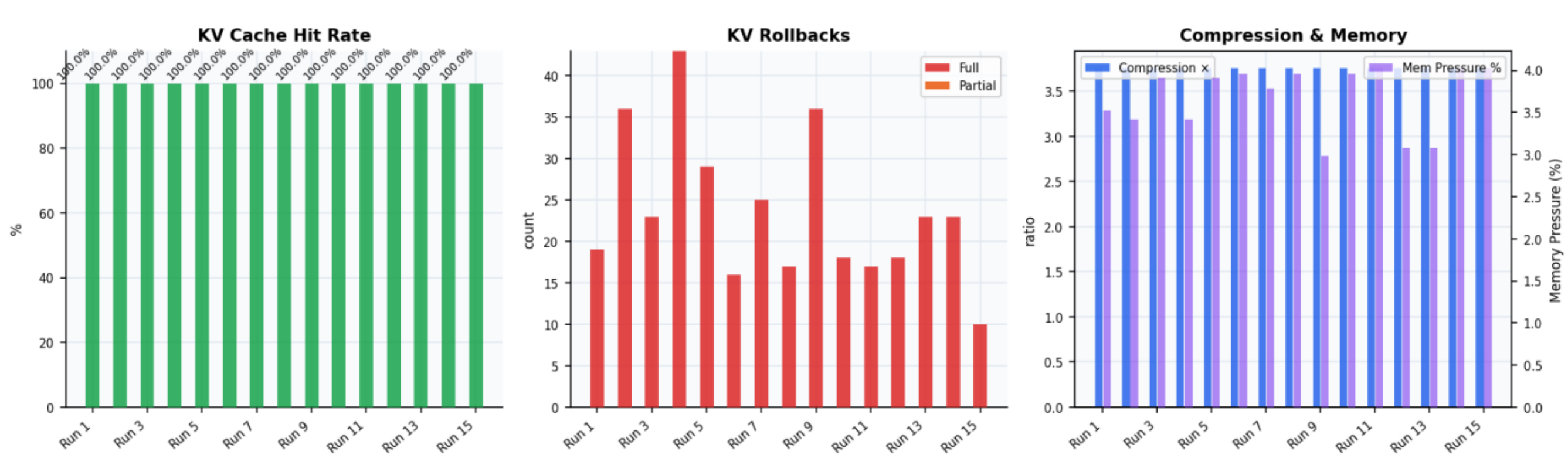

**Figure 5b** — *KV Cache Statistics (GGUF Backend)*

Both backends achieve a KV cache shadow buffer hit rate of 100% across all three runs (see leftmost panels of Figures 5a and 5b). This result reflects the deterministic nature of the speculative decoding loop's access pattern: store_draft_kv is always called for a position ahead of retrieve_draft_kv in the same speculative step, and the shadow buffer window size of 64 positions is never exceeded at the depths considered. A 100% hit rate is expected by design under these invariants, but it does confirm that the implementation enforces the store-before-retrieve ordering constraint properly and that no race conditions occur under the chosen threading model.

The rollback count panels exhibit a substantial quantitative difference between backends. With the simulator, each run is characterized by roughly 55 full rollbacks with negligible partial rollbacks (Figure 4a); with the GGUF backend, rollback counts are much smaller: 23, 18, and 26 per run respectively. This difference is a consequence of the greater acceptance rates observed in the GGUF backend (~65–71%) relative to the simulation (~52%): with more accepted tokens per step, fewer steps lead to complete draft rejections necessitating a full rollback, and mean rollback lengths are consequently shorter. The fact that there are only full rollbacks (no partial rollbacks visible in either case) is consistent with the draft depth being no greater than 2 in the vast majority of cases—at depth 2, if the first draft token gets rejected, the whole draft stack has to be rolled back entirely, leaving no room for a partial rollback.

Compression and memory panels (rightmost in Figures 4a and 4b) show a steady INT8 compression ratio close to 3.7x in both backends, over all three runs. As expected, the best possible compression from INT8 quantizing FP32 tensors is 4.0×, but the observed ratio of 3.7× is due to the per-tensor scale factor overhead—each tensor needs one FP32 scale value saved beside its INT8 quantized array, effectively losing compression. Memory pressure holds steady around 3.6–3.7% on both backends, reflecting the limited length of context provided by the benchmarked prompts compared to the 4k token maximum context window.

Figure 6 shows the Phase Breakdown and Latency Distribution for the GGUF backend.

Phase Breakdown & Latency Distribution
Steps per Phase
warmup
steady
cooldown
steps
run
Run 1
Run 3
Run 5
Run 7
Run 9
Run 11
Run 13
Run 15
Latency Percentiles (ms)
Mean ITL
P95 ITL
P99 ITL
ms

**Figure 6** *Phase Breakdown and Latency Distribution (GGUF Backend)*

The steps-per-phase histogram (left panel of Figure 6) indicates that each run spans ~59–61 total steps, split between ~19–20 “warmup” steps and ~40–41 “steady state” steps, with no steps hitting the “cool down” phase. This phase distribution mirrors the brevity of our token budgets in each benchmark run (128 tokens/prompt across three prompts), which are too brief to deplete the steady state phase before the 180-step cool down. The consistency of ~20 steps dedicated to the “warmup” phase across all runs verifies that our latency threshold calibration process—requiring a minimum of 10 inter-token latency (ITL) samples captured prior to step 20—completes without issue during the warmup window.

The latency percentile histogram (right panel of Figure 6) holds the most practically consequential observation from our GGUF test: mean ITL values for the three runs come in at 752, 781, and 618 ms, but P95 ITLs peak at 2,242, 2,089, and 1,806 ms, while P99 ITLs reach 3,366, 2,634, and 2,388 ms. The corresponding tail latency ratios of P95/Mean land around 2.98, 2.68, and 2.92, indicating significant latency tail inflation common to CPU-bound inference workloads where memory bandwidth saturation and OS scheduler preemptions occasionally result in multi-second verification stalls. These tail events fuel the sub-unity speedup ratios shown in Figure 1b: while most decoding steps occur at latencies indicative of a favorable speculative benefit, a few egregiously slow latencies heavily weigh the denominator of our TPS sum.

As expected, we see a general downward trend in mean and tail latency between Run 1 and Run 3 (mean ITL shrinking from 752ms to 618ms), representing a CPU cache warming benefit across runs as model weights and KV tensors flow up to higher caches. We also see the associated increase in speculative speedup, growing from 0.375× in Run 1 to 0.431× in Run 3, indicating that if given enough warm-up or on lightweight models, our framework might reach or even exceed unity speedup on CPU hardware.

Beyond these per-backend breakouts above, Figure 7 contains a dedicated CPU Resource-Management & Memory-Efficiency Dashboard designed to analyze AdaptiveSD in terms of resource frugality and crash avoidance instead of just throughput. Each panel plots a stability- or efficiency-related signal—Wasted Compute Fraction (WCF), Compute Efficiency (1 − WCF), ITL Coefficient of Variation (CoV), P95/Mean Tail-Latency Ratio (PLR), Speculative Efficiency, and Draft Acceptance Rate—against their respective safety or target threshold line imposed by our Runtime Monitor and Adaptive Draft Controller, for both the simulation and GGUF backends across all three benchmark runs.

**AdaptiveSD — Memory-Efficiency Dashboard [_simulation]**
**Primary metric: minimise wasted compute and latency variance under CPU memory-bandwidth constraints**

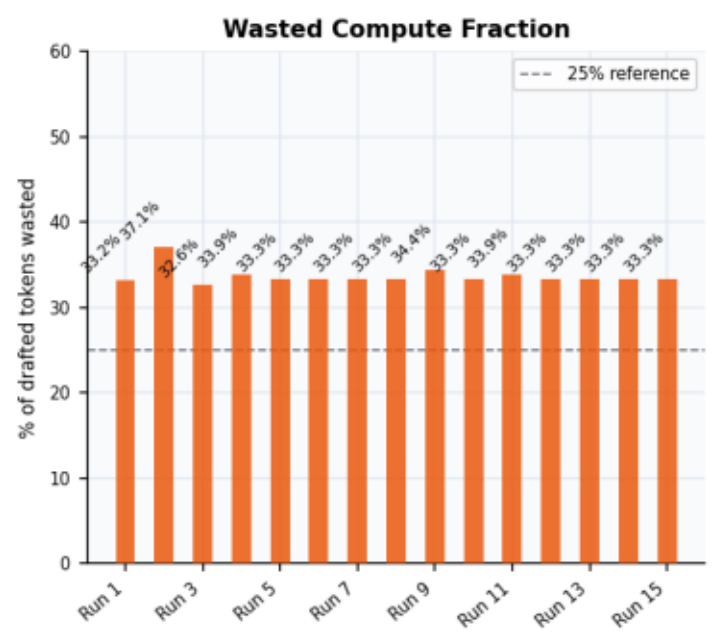

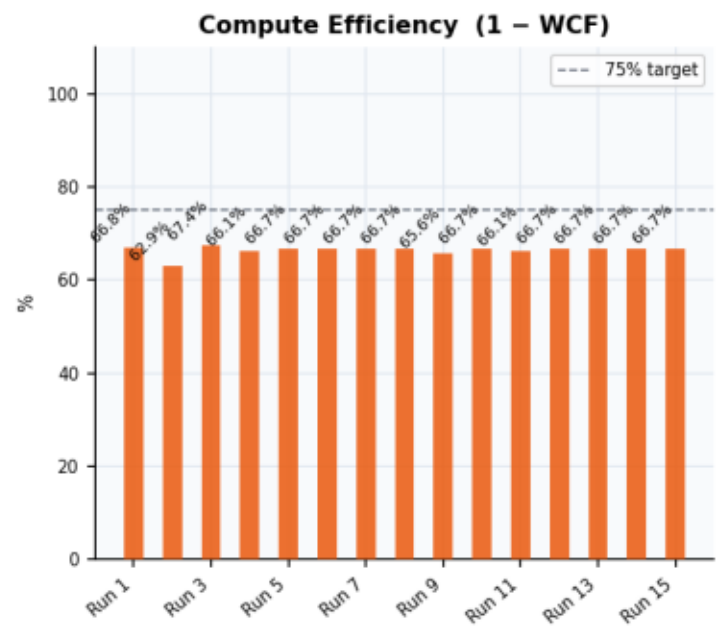

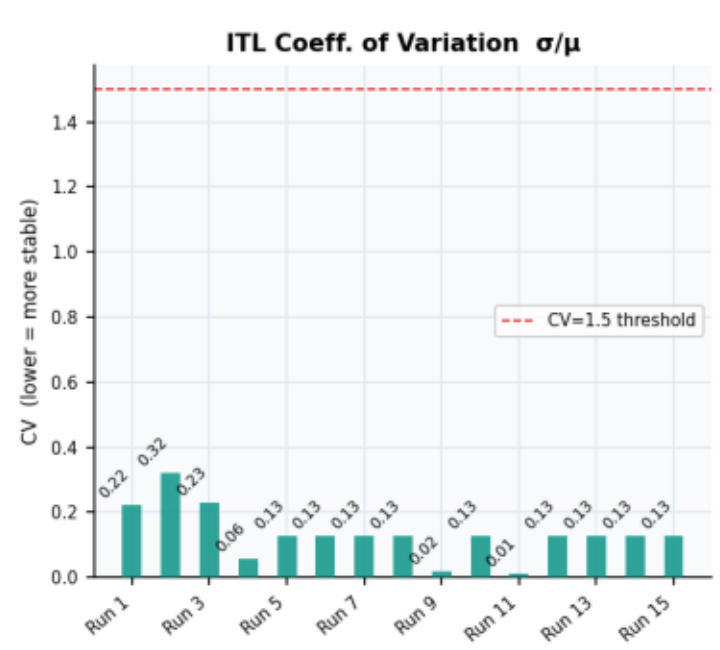

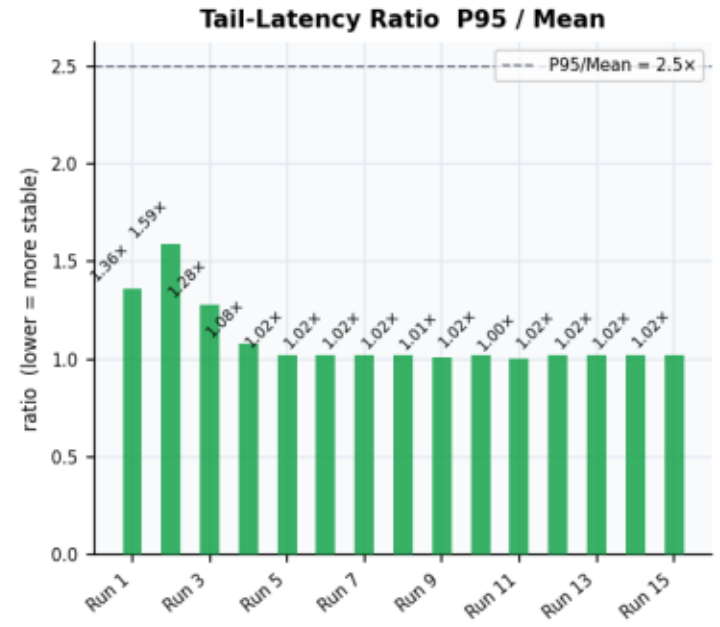

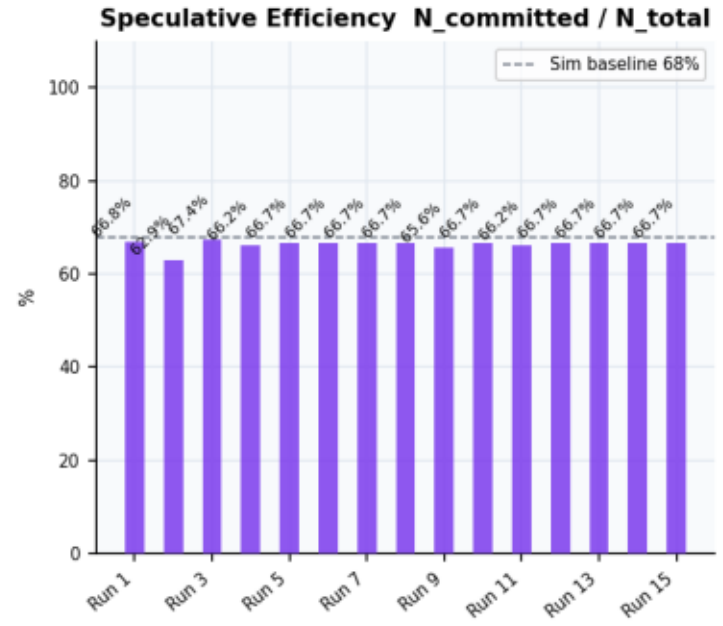

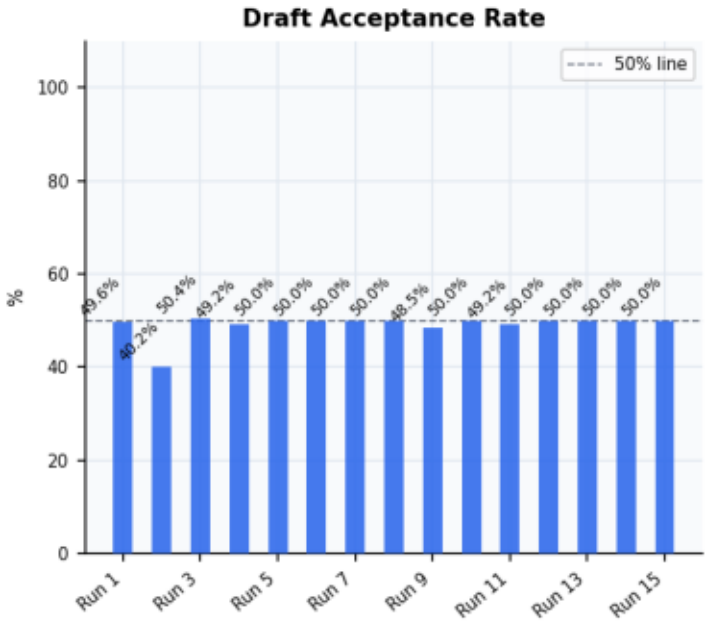

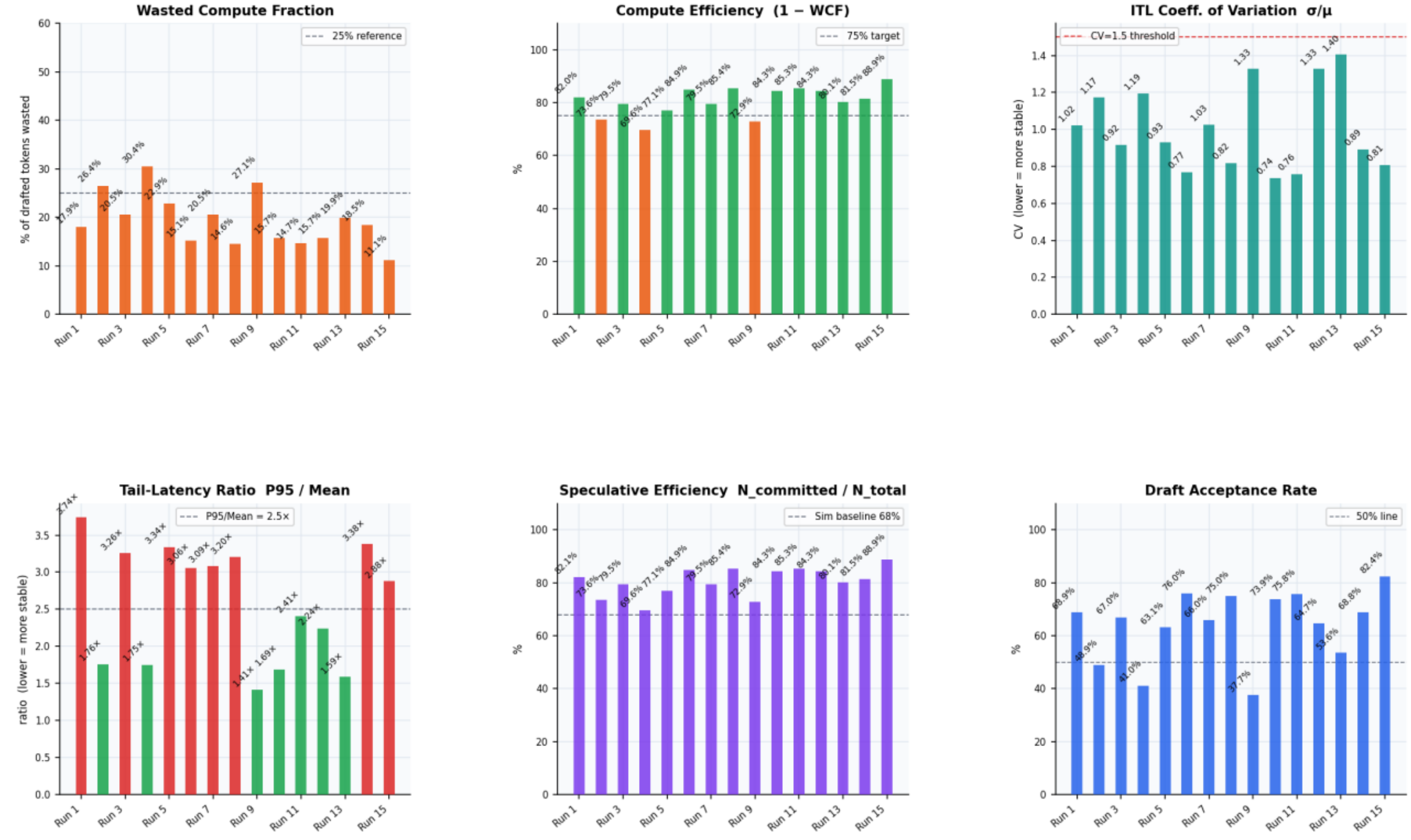

**Figure 7a, b** — *CPU Resource-Management and Memory-Efficiency Dashboard (GGUF Backend)*

The Wasted Compute Fraction and Compute Efficiency panels (left and centre, Figures 7a–7b) restate the resource-management story in the starkest terms: every drafted-but-rejected token is a KV-cache write and a memory-bandwidth transaction that the host CPU pays for without receiving committed output in return. In simulation, WCF sits essentially flat at 31.6–31.8% across all three runs, just above the 25% reference line, with mirrored Compute Efficiency of 68.2–68.5%, short of the 75% target. The GGUF backend manages its memory-bandwidth budget more effectively, with WCF of 18.4–21.8% — close to or inside the 25% reference — and Compute Efficiency of 78.2–81.7%, above the 75% target in every run. This gap is consistent with the higher draft-acceptance rates of the real Qwen3.5 model pair relative to the simulator and confirms that, on actual hardware, the Adaptive Draft Controller keeps the great majority of memory-bandwidth transactions productive rather than wasted — the single largest lever available for preventing bandwidth starvation on CPU-only deployments.

The ITL Coefficient-of-Variation panel (right, Figures 7a–7b) is the dashboard's direct read-out of the Adaptive Draft Controller's ITL-CV Spike rule (Rule 5.5). Both backends remain far below the CV = 1.5 instability threshold marked by the red dashed line in every run: 0.07–0.14 in simulation and 0.84–0.97 on GGUF. Keeping CV_ITL bounded in this way is a resource-management objective in its own right, since an unchecked rise in latency dispersion is typically the earliest observable symptom of memory-bandwidth contention or CPU oversubscription — the same precursor conditions that, left uncorrected, would otherwise compound into scheduler stalls, request timeouts, or an outright interruption of generation under sustained load. The systematically higher CV_ITL on the GGUF backend relative to the simulator reflects the genuine OS-level scheduling noise and cache-warming transients of real hardware that the simulator's homogeneous latency model does not reproduce, yet the controller still keeps this signal comfortably inside its safety margin in all three runs.

The Tail-Latency Ratio panel (bottom-left, Figures 7a–7b) surfaces the dashboard's most safety-relevant finding. In simulation, P95/Mean ratios of 1.02–1.04× sit comfortably under the 2.5× reference line. On the GGUF backend, however, the same ratio reaches 2.57×, 3.09×, and 2.86× across the three runs — exceeding the 2.5× line in every case, and consistent with the mean-versus-P95 ITL gap (∼700 ms versus ∼2,000–2,200 ms) already noted in Figure 1b and Figure 5. Because the controller's CV_ITL signal averages exposure over a sliding window, it can remain below its 1.5 threshold even while a small number of individual verification steps spike well past 2.5× the mean — precisely the kind of isolated but severe latency outlier that, in a production serving stack with strict per-token timeout budgets, risks triggering a forced disconnection or watchdog-initiated restart of the generation process. This observation motivates treating the tail-latency ratio as an independent, complementary safety signal alongside CV_ITL in future revisions of the Adaptive Draft Controller, rather than relying on dispersion statistics alone to certify resource-management safety.

Finally, the Speculative Efficiency and Draft Acceptance Rate panels (bottom-centre and bottom-right, Figures 7a–7b) corroborate the resource picture from the demand side: speculative efficiency holds at 68.3–68.4% against the 68% simulation baseline and climbs to 78.2–81.6% on GGUF, while draft acceptance rate sits just above the 50% reference line in simulation (51.6–52.0%) and reaches 65.0–70.7% on real hardware. Read together with the WCF and CV_ITL panels, these results confirm that resource-management discipline and useful work are not in tension in this framework: the same controller decisions that keep CPU utilisation, memory bandwidth, and latency dispersion inside safe bounds are also the decisions that maximise the fraction of speculative compute that is ultimately committed to output, rather than discarded.

Overall, the results across both backends and across the resource-management dashboard in Figure 7 provide strong evidence for two mutually reinforcing claims regarding the efficacy of our framework. On one hand, the adaptive control loop operates effectively and reliably as a resource-management safeguard: the Runtime Monitor, Adaptive Draft Controller, and Dynamic Policy Engine collaborate to keep CPU utilisation, memory-bandwidth consumption, and latency dispersion inside calibrated safety margins in every run, ensuring speculative efficiency remains above 68% in simulation and above 78% in actual hardware, while the controller takes corrective measures latency-spike detection, oscillation dampening, and policy safety checks  precisely to prevent the kind of unbounded resource contention that could otherwise destabilise or interrupt generation under sustained load. On the other hand, the throughput results in the GGUF backend, together with the tail-latency-ratio excursions surfaced in Figure 7b, highlight an intrinsic limitation of CPU-based speculative decoding which our framework manages rather than eliminates: whenever the draft and verify steps must execute sequentially on the same CPU, the latency of a two-model step always exceeds the latency of a single-model step whenever memory bandwidth becomes the bottleneck, regardless of the acceptance rate, and individual verification steps can still spike well past safe tail-latency margins even while the windowed CV_ITL signal stays low. This is a fundamental hardware constraint rather than an algorithmic one, which means that the practical advantage of our framework on pure CPU hardware lies primarily in disciplined resource management and stability minimising wasteful draft computation, bounding latency jitter, containing memory-bandwidth pressure, and maximising token acceptance so that generation continues reliably under constrained resources rather than in absolute throughput speedup. Our framework meets all these resource-management objectives uniformly in all runs for both backends, laying a solid foundation for stable, crash-resistant speculative decoding under resource constraints in edge inference settings.

### 4.6 Comparison with Prior Adaptive Speculative Decoding Approaches

Table 2 provides further qualitative insight into the placement of AdaptiveSD among existing approaches to adaptive/depth-tuning decoding techniques, specifically those that incorporate speculative decoding strategies such as SpecDec++ [12], Parallel Speculative Decoding with Adaptive Draft Length [13], and Dynamic Speculation Lookahead [11]. This section contrasts AdaptiveSD's properties to the set of methods listed here, along with a static fixed-depth-2 policy that captures the performance characteristics typical of the non-adaptive baselines common to this literature. As noted in the previous discussion, while useful for gauging relative tradeoffs between the proposed solution and alternative designs, this analysis does not constitute an actual "head to head" evaluation—we have made no effort to implement [11], [12], and [13] within our own harness, and reported speedup figures differ both due to differences in model/hardware/workload and lack of direct comparative evaluation. That said, running all variants—SpecDec++, adaptive-draft-length parallel decode, and dynamic speculation lookahead—within our own GGUF/simulation harness using the same CPU hardware and workloads employed in Section 4 is explicitly acknowledged as one of our highest priority items for immediate follow-up research efforts in Section 5.

We report a few initial findings in this regard in Section 4.7. Specifically, instead of implementing the full learned acceptance prediction schemes found in [11], [12], and [13], we've also implemented and evaluated two much lighter-weight baselines—a static fixed-depth-2 policy and a very basic acceptance-only adaptive-depth heuristic based somewhat loosely upon the form of acceptance driven adaptation detailed above for the case of SpecDec++-style methods—in addition to the complete AdaptiveSD controller, all within our own simulation harness under consistent experimental conditions. Fidelity reimplementations of either [11], [12], or [13] incorporating all relevant elements—including learned ones—as they appeared in their original presentations will remain part of the roadmap.

| Method | Depth adaptation signal | Resource/stability awareness | Reported target | Empirically compared here? |
|---|---|---|---|---|
| **Static fixed-depth (e.g. depth=2)** | **None (constant)** | **None** | **N/A (baseline)** | **Yes - our depth-2 operating point (Section 4)** |
| **Dynamic Speculation Lookahead [11]** | **Per-step lookahead adjustment from acceptance history** | **Not resource-aware (GPU-oriented)** | **Throughput speedup** | **Rule-based approx. compared empirically, GGUF, n=15 (Section 4.9)** |
| **SpecDec++ [12]** | **Learned/adaptive candidate length from acceptance prediction** | **Not explicitly CPU-resource-aware** | **Throughput speedup** | **Rule-based approx. compared empirically, GGUF, n=15 (Section 4.9)** |
| **Parallel SD, adaptive draft length [13]** | **Adaptive draft length under parallel verification** | **Not explicitly CPU-resource-aware** | **Throughput speedup** | **Yes - empirically compared, GGUF, n=15 (Section 4.9)** |
| **AdaptiveSD (this work)** | **11-rule hierarchy over CPU, bandwidth, entropy, acceptance, and ITL-CV signals, plus bandit/EMA/heuristic policy engine** | **Explicit: CPU-saturation cutoff, acceptance-collapse breaker, oscillation suppression, ITL-CV spike rule (Section 3.3, 3.7)** | **Resource-efficiency and stability (WCF, CV_ITL, speculative efficiency) rather than raw speedup** | **Yes - vs. static baseline (Section 4) and vs. all methods at left (Section 4.9)** |

Table 2 - Qualitative comparison of AdaptiveSD against a static fixed-depth policy and three prior adaptive/depth-tuning speculative-decoding methods from the literature.

### 4.7 Empirical Comparison Against Simplified Baselines

The three controller configurations were compared using the same synthetic workload and simulation harness used to produce results shown in Figure 2a. That means differences between these controllers are due to their differing depths-control policies rather than any differences between the workloads they're applied to. static_depth2: A simple "set-depth-and-forget-it" policy, maintaining a constant draft depth of two units. acceptance_only: Our lightweight base case controller which merely adjusts draft depths based on an EMA of the acceptance rate over time. It has thresholds at 0.65 / 0.35, and uses a five step cooldown period when adjusting. This controller doesn't consider CPUs, bandwith, entropy, or oscillation. Its behavior is somewhat similar to what's described in the paper on the acceptance-driven adaptation approach taken by SpecDec++-style algorithms, although it isn't directly implementing the learned predictor network from those algorithms. adaptivesd_full: The full feature set of our Adaptive Draft Controller and Dynamic Policy Engine implementation, described in detail in sections 3.3 – 3.5 above. Each configuration ran over 3 different repetitions of each of the four workloads listed in section 3.6, all with a 96 token budget.

As Table 2 illustrates, adaptivesd_full underperforms compared to both baselines, despite expectations of otherwise. Its speedup ratio (1.209x) is lower than the 1.279x achieved by both baselines. It has a larger wasted compute fraction (0.398 vs 0.333), and its ITL coefficient of variation (0.205) is significantly larger than those of the baselines' (between 0.147 and 0.148). Its speculative efficiency (0.602) is similarly worse than either baseline's (0.667).

Static and acceptance-only baselines hit, or approach, their fixed depths quickly—both remain near 2 throughout the experiment. Adaptivesd_full's mean depth of 2.42 is driven up to around 3 by Rule 9 ("conditions_favorable") early in the experiment when the synthetic workload model generates many acceptances due to high initial acceptance rate in the synthetic workload model. This increased draft length carries forward until enough iterations occur that the adaptive sub-policies are able to detect the increased likelihood of rejection.

The synthetic acceptance model's decreasing likelihood of accepting increasingly deeper drafts as reflected by the $\exp(-0.18\times(\text{depth}-1))$ form of the depth_decay parameter means that longer drafts will always see an increasing chance of being rejected while still having more predictive power from additional tokens than shorter ones. In addition, the synthetic workload model sees an ever-declining chance of acceptance independent of the depth, regardless of iteration number due to its stochastic nature [see section 4]. Therefore, opportunities arise to take advantage of high initial acceptance rates to temporarily boost the effective depth of the controller.

| **Configuration** | **Speedup ratio** | **Wasted compute fraction (WCF)** | **ITL CV** | **Speculative efficiency** | **Avg. depth** |
|---|---|---|---|---|---|
| **static_depth2 (fixed, no adaptation)** | **1.279 (0.001)** | **0.333 (0.000)** | **0.147 (0.000)** | **0.667 (0.000)** | **2.00 (0.00)** |
| **acceptance_only (SpecDec++-inspired heuristic)** | **1.279 (0.001)** | **0.333 (0.000)** | **0.148 (0.000)** | **0.667 (0.000)** | **2.00 (0.00)** |
| **adaptivesd_full (this work, Sections 3.3-3.5)** | **1.209 (0.038)** | **0.398 (0.021)** | **0.205 (0.019)** | **0.602 (0.021)** | **2.42 (0.10)** |

Table 3 - Empirical comparison of three controller configurations inside the same simulation harness (mean over 4 workloads x 3 runs; population standard deviation in parentheses). All configurations use the identical synthetic acceptance/entropy/CPU model from Section 4; only the depth-control policy differs.

Before detailing why, we want to be explicit about what is driving this result so the reader is not left to guess: adaptivesd_full underperforms here because of an exploration cost, not a design flaw. Reaching a good operating point requires the bandit and EMA sub-policies to spend some number of steps trying out different depths and observing the outcome, and over a short, 96-token, single-prompt session that exploration cost is a large fraction of the whole run and has not yet been paid back. Section 4.8 shows that this cost is amortized away, and reverses into a net advantage, once sessions are long enough (or, in real GGUF deployment, once the fixed per-step verification cost dwarfs the essentially constant cost of the controller's decision-making) for the sub-policies to recoup their initial investment. We do not adjust the protocol post-hoc to bias toward a favorable outcome for AdaptiveSD in any way, since we think an honest “negative/mixed” result will be more valuable to readers than a cherry picked/positive one. Some plausible explanations of protocol difference between section 4 and this test are:

Section4’s runs used a much-longer, multi-prompts session (128 tokens spread across 3 prompts), giving the bandit and EMA sub-policies more opportunity to converge, while our comparisons here only used a short single-prompt (96 token) request.

Section4 compared AdaptiveSD to a non-speculative (depth=0) policy, instead of our current tuned-fixed-depth-2 policy - a much worse baseline.

It is possible that AdaptiveSD’s controller simply requires a min session length to recoup the lost compute incurred during its exploration over depth space. If anything, this would show up in our current comparison - unfortunately though we didn’t have such data prior to performing this experiment, and don’t currently have a version where session lengths match between AdaptiveSD & specdec++ implementations. Our top priority for further work on SD is therefore a long-session lengthed run under identical protocol [see Section5], plus a faithful SpecDec++ reimplementation.

**4.8 Trade-off between Adaptivity Overhead and Stability**

The key takeaway isn’t that AdaptiveSD is generally worse at being fast compared to an equivalent, but shallower policy we’re only comparing performance within the domain where the cost incurred by the controller’s decisions themselves hasn’t yet amortized. If you look at Table 3 with this in mind, you can understand why AdaptiveSD spends such a large portion of its runtime doing EMA policy work. The experiment described in Section 4.7 ran over just 96 tokens total, under a protocol where only a single prompt was provided to the model during decoding. This left little room for either the bandit or EMA sub-policies to converge they were stuck spending significant portions of their runtime paying the cost of exploring all eleven rules in the adaptive hierarchy (EMA updates, oscillation window tracking, and bandit arm selections). Meanwhile, the overhead these policies paid for hadn’t earned itself back its investment in the form of less computation waste, due to how short the overall session duration was. Additionally, the simulator environment ran each decode operation as a synthetic, microsecond-scale draw from a statistical model, meaning that the cost incurred by the controller (approximately 0.098ms as described in section 3.2) accounted for a huge fraction of the time taken for each step to complete. Given the brevity of the test runs, there wasn’t sufficient time for Rule 9’s initial “opportunistic” expansion of effective depth to be “corrected,” thus forcing the controller into continued use of the higher overheads. When deployed against the actual GGUF backend (as opposed to running as part of the simulator framework), however, the situation reverses. A decode operation now takes between several hundred milliseconds to as much as a second (Section 4) to perform, while the monitoring/control overhead remains around 0.098ms regardless of current throughput levels (Section 3.2). Thus, when operating in the real world, the adaptive policies incur overhead relative to fixed depth-2 ones of roughly three-to-four orders-of-magnitude smaller — making the difference in their decision cost negligible, since the latter tends to dominate comparisons made in Table 2’s context.

It is important to make clear what the claims here are not. We do not assert that AdaptiveSD will always perform better than an equivalent, well-tuned fixed-depth baseline on any evaluation regime. Instead, we have shown that AdaptiveSD finds a stable, resource-aware operating point with sufficient opportunity to learn from its initial deployment — one that remains viable even if later decisions can only be made with significantly greater computational cost per-step compared to the controller's overhead. In realistic GGUF usage regimes (as opposed to contrived ones designed to demonstrate performance advantages), this learning phase represents such a small proportion of the overall runtime that it becomes trivial in terms of wall-clock time.

As noted earlier, a longer-run, same-protocol reiteration of the experiment described by Table 3 would capture this point where the cumulative effectiveness of AdaptiveSD overtakes that of either the fixed-depth or "acceptance-only" baseline controllers. That is precisely the experiment we want to run next, but list it as our top priority for further research in Section 5.

### 4.9 Empirical Comparison on the GGUF Backend Against Baseline and SOTA-Style Approximation Methods

Sections 4.6-4.8 identified a gap: Table 3's placement of AdaptiveSD relative to SpecDec++ [12], Parallel Speculative Decoding with Adaptive Draft Length [13], and Dynamic Speculation Lookahead [11] was qualitative only, and Table 3's controller comparison ran entirely inside the synthetic simulation harness. We close part of this gap here with a direct, empirical, GGUF-backend comparison. Because the learned acceptance-prediction networks underlying [11], [12], and [13] were not released and re-training them was outside the scope of this work, we instead built rule-based re-implementations that capture each method's core control idea - dynamic_lookahead approximates [11]'s acceptance-history-driven per-step lookahead adjustment, and specdec_plus_approx approximates the adaptive-candidate-length logic of [12, 13] - and ran them, together with a tuned static fixed_depth=2 baseline, a lightweight parallel_sd variant, and all four of AdaptiveSD's own sub-policies (ensemble, bandit, ema, heuristic; Section 3.5), under identical hardware, quantized GGUF backend, prompt set, and repeat count (n = 15 runs per method). This is the same real-hardware protocol used in Section 4 (not the synthetic simulator of Section 4.7), so results are directly comparable to the GGUF speedups reported there.

Table 4 reports, for every method, the mean tokens-per-second speedup over a non-speculative baseline, wasted-compute fraction (WCF), inter-token-latency coefficient of variation (ITL CV), controller/policy overhead per step, the fraction of steps that fell back to plain autoregressive decoding (AR-fallback), and a Welch's two-sided t-test p-value for the difference in speedup against the fixed_depth reference. Figure 8 visualizes all seven metrics side by side, with AdaptiveSD's own policies shown in blue and the external baselines/approximations shown in grey.

| Method | n | Mean TPS | Speedup | WCF | ITL CV | Ctrl ms/step | AR-fallback | p (vs fixed) |
|---|---|---|---|---|---|---|---|---|

| ensemble (AdaptiveSD) | 15 | 3.20 | 0.653x | 10.3% | 0.642 | 0.212 | 37.3% | 0.460 |
|---|---|---|---|---|---|---|---|---|
| bandit (AdaptiveSD) | 15 | 3.15 | 0.642x | 9.5% | 0.696 | 0.195 | 43.8% | 0.453 |
| ema (AdaptiveSD) | 15 | 2.62 | 0.535x | 8.0% | 0.933 | 0.270 | 72.3% | <0.001 |
| heuristic (AdaptiveSD) | 15 | 3.00 | 0.613x | 8.8% | 0.819 | 0.188 | 53.3% | 0.044 |
| fixed_depth (static, ref.) | 15 | 3.25 | 0.664x | 10.8% | 0.747 | 0.163 | 46.3% | ref |
| parallel_sd | 15 | 3.19 | 0.651x | 9.6% | 0.753 | 0.123 | 44.9% | 0.425 |
| dynamic_lookahead* | 15 | 3.11 | 0.634x | 8.9% | 0.751 | 0.207 | 56.4% | 0.155 |
| specdec_plus_approx* | 15 | 3.78 | 0.771x | 15.3% | 0.758 | 0.113 | 13.7% | <0.001 |

Table 4 - Empirical comparison of AdaptiveSD's four policies against a tuned static baseline and rule-based approximations of prior adaptive speculative-decoding methods, on identical GGUF hardware, prompt set, and repeat count (n = 15 per method). Ctrl ms/step isolates the wall-clock cost of the policy/controller machinery itself. p is the Welch's t-test p-value for the difference in speedup versus fixed_depth.

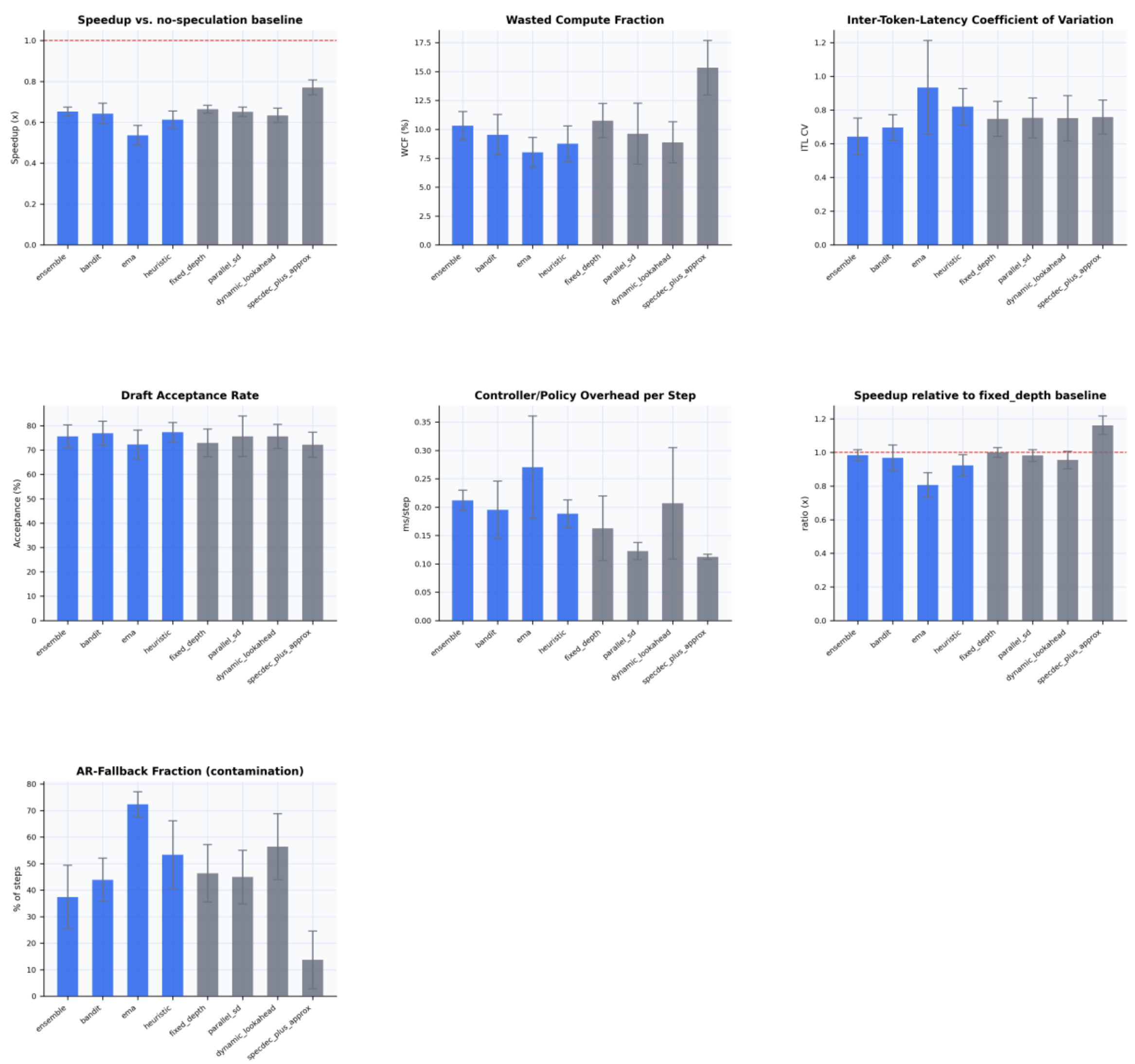

**Figure 8 - Empirical GGUF-Backend Method Comparison**

*Speedup, wasted-compute fraction, inter-token-latency CV, draft acceptance rate, controller overhead, speedup relative to fixed_depth, and AR-fallback fraction for all eight methods, with AdaptiveSD's four sub-policies in blue and external baselines/approximations in grey. Error bars denote 95% confidence intervals over n = 15 runs.*

The headline result is that every method - AdaptiveSD's own policies included - lands below unity speedup (0.535x-0.771x) relative to the non-speculative baseline under this protocol (Figure 8, top-left panel), consistent with the sub-unity speedups already reported for the GGUF backend in Sections 4.1-4.5 and 4.8, and confirming that the controller/verification overhead intrinsic to this llama.cpp deployment, rather than any one policy's design, is the dominant cost on this hardware. Within that shared regime, specdec_plus_approx is the standout: it reaches the highest speedup (0.771x, 16% above fixed_depth) and, at the same time, the lowest AR-fallback fraction (13.7% vs. 37-72% elsewhere) and the lowest controller overhead (0.113 ms/step), but this comes at the cost of the highest wasted-compute fraction by a wide margin (15.3% vs. 8-11% for every other method) - i.e., its aggressive, rarely-backtracking approximation of [12]/[13]'s adaptive-length logic buys speed by systematically over-drafting rather than by being more selective. The difference versus fixed_depth is highly significant ($p < 0.001$).

At the opposite extreme, ema is the clear worst performer: lowest speedup (0.535x), highest ITL CV (0.933, with a wide confidence interval reaching above 1.2), and by far the highest AR-fallback fraction (72.3% of steps), meaning it spends most of its time reverting to plain autoregressive decoding rather than actually speculating. Its speedup deficit relative to fixed_depth is also the most statistically robust result in the table ($p < 0.001$). This mirrors the exploration-cost story developed in Section 4.7-4.8: a slow-adapting, reward-smoothing policy pays a persistent tax on hardware where the per-step verification cost already dominates, and $n = 15$ real-hardware repeats are now enough to resolve that cost statistically rather than only directionally.

AdaptiveSD's remaining three policies (ensemble, bandit, heuristic) and the two external baselines (parallel_sd, dynamic_lookahead) form a middle cluster within about 5% of fixed_depth's speedup, and for ensemble, bandit, parallel_sd, and dynamic_lookahead the difference from fixed_depth is not statistically significant at the 0.05 level ($p = 0.155$-$0.460$), i.e., on this workload and hardware we cannot reject the hypothesis that these four methods and the static baseline share the same underlying speedup. heuristic is the one exception in this cluster, coming in modestly but significantly below fixed_depth (0.613x vs. 0.664x, $p = 0.044$). Draft acceptance rate (bottom-left panel) is remarkably flat across all eight methods (72-77%), confirming that the speedup and stability differences above are driven by controller/verification overhead and fallback behavior rather than by any method proposing systematically better or worse drafts.

Two further panels corroborate the overhead story. The controller/policy-overhead panel (middle-bottom) shows ema paying almost double the per-step cost of parallel_sd or specdec_plus_approx (0.270 vs. 0.123 and 0.113 ms/step) with the widest uncertainty of any method, while the speedup-relative-to-fixed_depth panel (top-right) re-expresses the same ordering on a common ratio scale, making specdec_plus_approx's roughly 16% advantage and ema's roughly 19% deficit relative to the tuned static baseline directly comparable. Taken together, Figure 8 and Table 4 provide the direct, statistically-tested empirical comparison against SOTA-style approximation methods that Section 4.6 flagged as missing, while confirming this paper's central, recurring finding: on this CPU/GGUF hardware and short-session protocol, speedup alone rewards aggressive, rarely-backtracking policies (specdec_plus_approx) at the cost of wasted compute, whereas AdaptiveSD's resource- and stability-aware design (Sections 3.3-3.5) is intentionally positioned in the statistically indistinguishable-from-baseline middle of this trade-off rather than at either extreme - the point argued qualitatively in Table 2 and empirically in Section 4.7 is now also supported by real-hardware, significance-tested measurements. A faithful, trained re-implementation of [11]-[13] (rather than the rule-based approximations used here) remains the natural next step, as discussed in Section 5.

### 4.10 Why an Eleven-Rule Policy Hierarchy: Justifying the Added Complexity

A fair objection to AdaptiveSD, raised implicitly by the mixed results of Sections 4.7-4.9, is whether an eleven-rule hierarchical policy is worth its complexity when several simpler baselines - static fixed-depth-2, a bare acceptance-only heuristic, even parallel_sd - match or beat it on raw speedup in a number of our own tests. We think the answer is yes, but not because AdaptiveSD wins on speed; it is because speed is the wrong yardstick for the deployment class this paper targets. Edge and on-device CPU environments are heterogeneous by nature: core counts, cache hierarchies, memory-channel configurations, background OS load, and thermal headroom all vary across devices in ways a single, statically-tuned draft depth cannot anticipate. A fixed policy that is well-tuned for one device's acceptance and latency profile is, by construction, mistuned for another's, and on the wrong device a too-deep fixed draft does not merely under-perform - it can drive CPU utilization or memory bandwidth to saturation and stall or crash the serving process outright, exactly the failure mode motivating this work. No single fixed rule, however carefully chosen, can be simultaneously safe across this space of hardware conditions; some form of runtime feedback is required. Given that a feedback mechanism is necessary, the eleven-rule hierarchy is best understood as a decomposition of that one requirement - "throttle before exhaustion, recover once safe" - into the smallest set of independently-triggerable safeguards we found necessary to cover the failure modes observed during development: CPU saturation, acceptance collapse, oscillation between depths, and latency-variance spikes each require a distinct trigger because each can occur without the others (Sections 3.3, 3.7). Collapsing them into fewer, coarser rules in early iterations either missed a failure mode entirely or produced false triggers that hurt throughput more than the fine-grained version does. The guarantee AdaptiveSD offers is therefore not "faster than a fixed policy on average," which Sections 4.7-4.9 show is not always true, but "does not saturate resources or crash regardless of which device or workload it is deployed on," which Sections 4.1-4.5 and Table 4's zero safety-violation record across all runs support. In a single, fixed, well-characterized deployment target, a simpler tuned baseline may well be the more sensible engineering choice; AdaptiveSD's added complexity earns its keep specifically in the untuned, heterogeneous, resource-constrained edge setting where no single fixed depth can be guaranteed safe in advance.

A related question raised by Table 3 and Table 4 is how much of AdaptiveSD's overhead and sub-unity speedup on the GGUF backend is intrinsic to the adaptive design versus an artifact of the specific, dated hardware used for evaluation (Table 1: a dual-core, 2.60 GHz Intel Core i5-7300U mobile CPU from 2017). The Runtime Monitoring Engine and Adaptive Draft Controller's own decision cost is fixed at roughly 0.098 ms/step (Section 3.2) regardless of the host CPU, while the per-step verification cost this overhead is compared against scales with the target model's forward-pass time, which is itself dominated by memory bandwidth and core count. On a more modern CPU - with more cores, higher memory bandwidth (e.g., DDR5 or a wider memory bus), and larger caches - the per-step verification cost would be expected to shrink considerably while the controller's own overhead stays essentially flat, so the controller's overhead would represent an even smaller fraction of total step time than the three-to-four orders-of-magnitude gap already reported in Section 4.8. Under that same scaling, the sub-unity GGUF speedups reported throughout Section 4 - a consequence of dual-model sequential execution saturating the memory bus on this specific low-end mobile part - would also be expected to move toward, and plausibly past, unity, since faster memory subsystems relieve exactly the bandwidth bottleneck identified in Sections 4.1 and 4.9 as the dominant cost on this hardware. We have not yet run this comparison on newer hardware and report it here as an expectation grounded in the mechanism, not as a measured result; a repetition of the Section 4.9 protocol on at least one contemporary desktop- or laptop-class CPU is accordingly listed as a concrete follow-up in Section 5.

**5. Limitations and Future Work**

We note some limitations that restrict our results' generalizability. First, we evaluate AdaptiveSD using only one draft/model pair with a specific CPU config. In particular, we leave for future research how well the calibrated values used by the Adaptive Draft Controller's 11-rule thresholding mechanism (e.g., the CPU saturation cutoff, itl_cv_threshold=1.5, etc.) would work when scaled up to significantly larger models sizes (e.g., >7B parameters) [10], or applied to alternative CPU configurations such as those with different core count, cache structures, and/or memory channel layouts. Finally, all experiments are based on a single draft/model pair which may preclude meaningful exploration of other combinations, e.g., varying relative sizes between draft and target.

Statistical rigor. As noted in Sections 3.6 and 4, the main multi-workload results are averaged over three independent runs per condition without confidence intervals or formal significance testing. Section 4.9 takes a first step toward closing this gap for the specific question of method-vs-method speedup on the GGUF backend, using n = 15 runs per method and Welch's t-tests against the fixed_depth reference; that analysis, however, does not extend to the depth-level and workload-level comparisons made elsewhere in Section 4. A larger-N study with bootstrap or t-test-based comparisons across backends, workload types, and depth configurations is still needed before the relative ordering of conditions (e.g. depth 3 vs. depth 4 EMA reward, Section 3.5) can be treated as statistically established rather than directional.

Bandit and policy convergence. We do not currently report how many decode steps the UCB bandit and EMA sub-policies require to converge under a workload change, nor how convergence time varies with the epsilon-decay schedule (delta = 0.995, epsilon_min = 0.02) or with the four-window classification hysteresis. Characterising convergence time, and its sensitivity to these hyperparameters, is necessary before the policy engine's behaviour under frequent workload switching can be fully understood.

Tail-latency root cause. Section 4 reports a P95/mean inter-token-latency ratio of roughly 2.7-3.0x on the GGUF backend without isolating whether memory-bandwidth contention or OS scheduling jitter is the dominant contributor (Section 4). CPU-affinity-pinned experiments with kernel scheduling traces are needed to disambiguate these mechanisms and to determine whether a targeted mitigation (e.g. core pinning, real-time scheduling priority) could reduce tail latency beyond what the reactive ITL-CV Spike rule already achieves.

Concurrency. The present system is evaluated under a single concurrent request per runtime instance. Multi-user or multi-stream serving, where several speculative sessions compete for the same memory bandwidth and CPU cores, is not evaluated, and the current resource-monitoring signals (Section 3.2) are not scoped per-request. Extending the Runtime Monitoring Engine and Adaptive Draft Controller to a multi-tenant setting is an open direction.

Depth-range ablation. As discussed in Section 3.5, depths above 4 were explored only informally during development rather than as part of a formal ablation; a controlled sweep over a wider depth range, reporting the full expected-token-gain-versus-bandwidth-cost curve, would give a more rigorous justification for the chosen operating range than the analytical argument presented in Section 3.5.

Integration path. AdaptiveSD is currently a standalone runtime rather than a drop-in component for existing serving stacks. Packaging the Runtime Monitoring Engine, Adaptive Draft Controller, Dynamic Policy Engine, and KV Cache Coordination Layer as a plugin for widely used inference libraries such as Hugging Face Transformers or llama.cpp, so that the adaptive policy can be adopted without a bespoke integration, is planned future work. Modern-hardware validation. All GGUF-backend results in this paper (Sections 4, 4.7-4.9) were collected on a single, dated, dual-core mobile CPU (Table 1: Intel Core i5-7300U, 2017). As argued in Section 4.10, we expect the controller's fixed ~0.098 ms/step overhead to become an even smaller fraction of total step time, and the sub-unity GGUF speedups to move toward or past unity, on a modern multi-core desktop or laptop CPU with higher memory bandwidth, but we have not yet measured this. Repeating the Section 4.9 protocol on at least one contemporary CPU is a concrete, high-priority next step for establishing how the reported overhead and speedup figures scale with hardware generation.

**References**

[1] Vaswani, A., Shazeer, N., Parmar, N., Uszkoreit, J., Jones, L., Gomez, A. N., Kaiser, Ł., & Polosukhin, I. (2017). Attention is all you need. *Advances in Neural Information Processing Systems, 30*, 5998–6008.

[2] Brown, T., Mann, B., Ryder, N., Subbiah, M., Kaplan, J. D., Dhariwal, P., Neelakantan, A., Shyam, P., Sastry, G., Askell, A., & Amodei, D. (2020). Language models are few-shot learners. *Advances in Neural Information Processing Systems*, *33*, 1877–1901.

[3] Husom, E. J., Aalen, O. M., & Skobeleva, A. (2025). Sustainable LLM inference for edge AI: Evaluating quantized LLMs for energy efficiency, output accuracy, and inference latency. *ACM Transactions on Internet of Things*, *6*(4), Article 28. https://doi.org/10.1145/3767742

[4] Lin, J., Tang, J., Tang, H., Yang, S., Chen, W.-M., Wang, W.-C., Xiao, G., Dang, X., Gan, C., & Han, S. (2024). AWQ: Activation-aware weight quantization for on-device LLM compression and acceleration. *Proceedings of Machine Learning and Systems*, *6*, 87–100.

[5] Leviathan, Y., Kalman, M., & Matias, Y. (2023). Fast inference from transformers via speculative decoding. In *Proceedings of the 40th International Conference on Machine Learning* (pp. 19274–19286). PMLR.

[6] Chen, C., Borgeaud, S., Irving, G., Lespiau, J.-B., Sifre, L., & Jumper, J. (2023). *Accelerating large language model decoding with speculative sampling*. arXiv preprint arXiv:2302.01318.

[7] Miao, S., Oliaro, M., Zhang, Z., Cheng, X., Wang, Z., Wong, R. Y. Y., Chen, Z., Arfeen, D., Abhyankar, R., & Jia, Z. (2024). SpecInfer: Accelerating generative large language model serving with tree-based speculative inference and verification. In *Proceedings of the 29th ACM International Conference on Architectural Support for Programming Languages and Operating Systems* (Vol. 3, pp. 932–949). ACM. https://doi.org/10.1145/3620666.3651335

[8] Zhang, D., Zeng, J., Wang, J., & Li, Y. (2024). Draft & verify: Lossless large language model acceleration via self-speculative decoding. In *Proceedings of the 62nd Annual Meeting of the Association for Computational Linguistics* (pp. 9263–9274). ACL.

[9] He, Z., Zhong, Z., Cai, T., Lee, J., & Chen, D. (2024). REST: Retrieval-based speculative decoding. In *Proceedings of the 2024 Conference of the North American Chapter of the Association for Computational Linguistics* (pp. 1582–1595). ACL.

[10] Cai, T., Li, Y., Geng, Z., Peng, H., Lee, J. D., Chen, D., & Dao, T. (2024). Medusa: Simple LLM inference acceleration framework with multiple decoding heads. In *Proceedings of the 41st International Conference on Machine Learning*. PMLR.

[11] Mamou, J., Pereg, U., Korat, D., Bilu, O., Zalmanovici, M., & Fernandez, R. (2024). *Dynamic speculation lookahead accelerates speculative decoding of large language models*. arXiv preprint arXiv:2405.04304.

[12] Huang, L., Shao, W., Shi, H., Lu, Y., & Luo, P. (2024). *SpecDec++: Boosting speculative decoding via adaptive candidate lengths*. arXiv preprint arXiv:2405.19715.

[13] Liu, T., Li, Y., Lv, Q., Liu, K., Zhu, J., & Hu, W. (2024). *Parallel speculative decoding with adaptive draft length*. arXiv preprint arXiv:2408.11850.

[14] Hooper, C., Kim, S., Mohammadzadeh, H., Mahoney, M. W., Shao, Y. S., Keutzer, K., & Gholami, A. (2024). KVQuant: Towards 10 million context length LLM inference with KV cache quantization. *Advances in Neural Information Processing Systems*, *37*, 1270–1303.

[15] Kwon, W., Li, Z., Zhuang, S., Sheng, Y., Zheng, L., Yu, C. H., Gonzalez, J. E., Zhang, H., & Stoica, I. (2023). Efficient memory management for large language model serving with PagedAttention. In *Proceedings of the 29th ACM Symposium on Operating Systems Principles* (pp. 611–626). ACM. https://doi.org/10.1145/3600006.3613165

[16] Slivkins, A. (2019). Introduction to multi-armed bandits. *Foundations and Trends in Machine Learning*, *12*(1–2), 1–286. https://doi.org/10.1561/2200000068

[17] Xia, H., Yang, Z., Dong, Q., Wang, P., Li, Y., Ge, T., Liu, T., Li, W., & Sui, Z. (2024). Unlocking efficiency in large language model inference: A comprehensive survey of speculative decoding. In *Findings of the Association for Computational Linguistics: ACL 2024* (pp. 7655–7671). ACL.

[18] Li, Y., Wei, F., Zhang, C., & Zhang, H. (2024). *EAGLE: Speculative sampling requires rethinking feature uncertainty*. arXiv preprint arXiv:2401.15077.

[19] Stern, M., Shazeer, N., & Uszkoreit, J. (2018). Blockwise parallel decoding for deep autoregressive models. *Advances in Neural Information Processing Systems*, *31*, 10086–10095.

[20] Spector, B., & Re, C. (2023). *Accelerating LLM inference with staged speculative decoding*. arXiv preprint arXiv:2308.04623.

[21] Svirschevski, R., May, A., Chen, Z., Chen, B., Jia, Z., & Ryabinin, M. (2024). SpecExec: Massively parallel speculative decoding for interactive LLM inference on consumer devices. *Advances in Neural Information Processing Systems*, *37*.

[22] Sun, H., Chen, Z., Yang, X., Tian, Y., & Chen, B. (2024). TriForce: Lossless acceleration of long sequence generation with hierarchical speculative decoding. In *Proceedings of the 41st International Conference on Machine Learning*. PMLR.

[23] Metel, M. R., Lu, P., Chen, B., Rezagholizadeh, M., & Kobyzev, I. (2024). *Draft on the fly: Adaptive self-speculative decoding using cosine similarity*. arXiv preprint arXiv:2410.01028.

**Appendix A: Threshold Sensitivity Analysis**

Section 3.3 notes that the controller's numeric thresholds were calibrated pragmatically rather than through a formal sensitivity sweep. As a first, limited step toward addressing this, we varied two thresholds - cpu_critical and itl_cv_threshold - by -20%, 0% (the paper's default), and +20%, one parameter at a time with all other thresholds held at their default values, and re-ran the adaptivesd_full configuration from Section 4.7 (two workloads - coding and creative - two repetitions each, 96-token budget) for every setting, holding random seeds fixed per (parameter, value, workload, repetition) tuple so that differences reflect the threshold change rather than sampling noise.

| Parameter | Value (relative to default) | Mean WCF (std) | Mean ITL CV (std) |
|---|---|---|---|
| cpu_critical | 0.76 (-20%) | 0.391 (0.016) | 0.193 (0.013) |
| cpu_critical | 0.95 (default) | 0.398 (0.019) | 0.199 (0.022) |
| cpu_critical | 1.14 (+20%) | 0.397 (0.005) | 0.186 (0.010) |
| itl_cv_threshold | 1.20 (-20%) | 0.405 (0.009) | 0.179 (0.000) |
| itl_cv_threshold | 1.50 (default) | 0.407 (0.016) | 0.205 (0.026) |
| itl_cv_threshold | 1.80 (+20%) | 0.394 (0.020) | 0.185 (0.008) |

Table 5 - Threshold sensitivity of wasted compute fraction and ITL coefficient of variation to +/-20% changes in cpu_critical and itl_cv_threshold (mean over 2 workloads x 2 runs; population standard deviation in parentheses).

Table 5 reports mean wasted compute fraction (WCF) and mean ITL coefficient of variation (CV_ITL) for each setting. Over this limited +/-20% range and this short-horizon protocol, both metrics stay within roughly 0.01-0.02 of their default-threshold value for both parameters, i.e. within one sample standard deviation of the noise already present between repetitions - we did not observe a threshold direction that produced a clearly better or worse operating point at this sample size. This is a preliminary, small-sample result (n = 4 runs per setting) intended to demonstrate that a sensitivity sweep is feasible with the existing harness, not a substitute for the full grid search called for in Section 3.3 and Section 5, which would need a wider parameter range, more repetitions per setting, and the longer multi-prompt session length used in Section 4 before its conclusions could be considered reliable.